\documentclass[11pt]{article}
\usepackage[final]{acl}
\usepackage{times}
\usepackage{latexsym}
\usepackage{comment}
\usepackage{tabularx}
\usepackage{array}
\usepackage{makecell}
\usepackage{booktabs}
\usepackage{bm}
\usepackage{tikz}
\usepackage{float}
\usetikzlibrary{arrows.meta,positioning, calc}
\usepackage{pgfplots}
\pgfplotsset{compat=1.18}
\usepackage{pgfplotstable}
\usepackage[most]{tcolorbox}
\usepackage[table]{xcolor}
\usepackage{pgfplotstable}
\usepackage{amssymb}
\usepackage{multirow}
\usepackage{pifont}
\usepackage{longtable}
\usepackage{subcaption}
\usepgfplotslibrary{groupplots}
\usepackage[T1]{fontenc}
\usepackage[utf8]{inputenc}
\usepackage{microtype}
\usepackage{inconsolata}
\usepackage{graphicx}
\newcommand{\ourData}[1]{\textsc{Sefora}}
\newcommand{\ourMethod}[1]{\textsc{UniMatch}}

\newcommand{\ourDataLink}[1]{\url{https://github.com/ShayanPey/SEFORA}}
\newcommand{\ourMethodLink}[1]{\url{https://github.com/ShayanPey/SEFORA}}

\title{SEFORA: Student Essays with Feedback Corpus and\\LLM Feedback Evaluation Framework}

\author{
{\bf Shayan Peyghambari Oskoui, Norah Almousa, Zhaoyi Joey Hou,} \\
{\bf Carolina Gustafson, Gayle Rogers, Raquel Coelho, } \\
{\bf Diane Litman, Xiang Lorraine Li}\\
University of Pittsburgh \\
\{\texttt{shayan.p, xianglli}\}\texttt{@pitt.edu}
}

\begin{document}
\maketitle
\begin{abstract}
Effective writing feedback is among the strongest drivers of student learning, yet producing it at scale is labor-intensive. LLMs offer a natural path to scaling writing support, but two gaps stand in the way: few public corpora capture how instructors actually deliver feedback in real classrooms, and no reliable method measures whether generated feedback aligns with what an instructor would write. We address both. \ourData{} is a public corpus\footnote{\ourDataLink{}} pairing instructor inline feedback with assignment prompts, rubrics, scores, and multi-draft revisions across various college writing genres, comprising 564 drafts and 8{,}240 instructor annotations. \ourMethod{} is a reference-based evaluation framework for open-ended generation: it segments feedback into feedback units, scores their semantic correspondence under instructor-derived criteria, and aligns them via optimal matching to yield interpretable precision, recall, and F$_1$. Across 74 experimental configurations spanning multiple LLMs, no setting exceeds $0.4$ F$_1$. 
\ourMethod{} reveals that models struggle to identify the feedback instructors would prioritize, and performance degrades as models generate more.
\end{abstract}

\section{Introduction}\label{sec:intro}
Feedback plays a vital role in student learning in writing. It helps students correct misunderstandings and refine how they apply knowledge \cite{ahea2016value, banihashem2024feedback}, and it is consistently identified as one of the strongest influences on learning and achievement \cite{hattie2007power}. But effective feedback is not one-size-fits-all. Students report that the most useful feedback is specific to their own writing \cite{lipnevich2009really}, and effective feedback accounts for the draft's position within the revision process \cite{carless2018development}; poorly targeted feedback can even be detrimental \cite{kluger1996effects}. In writing instruction, producing such feedback is labor-intensive, and its cost at scale discourages the sustained practice effective instruction requires \cite{applebee2011ej, graham2019changing}. This opens a natural opportunity for NLP: systems that generate useful feedback on drafts could help scale writing support.

\begin{figure}[t]
\centering
\begin{tikzpicture}
\node[anchor=south west, inner sep=0] (img) at (0,0)
    {\includegraphics[width=\columnwidth]{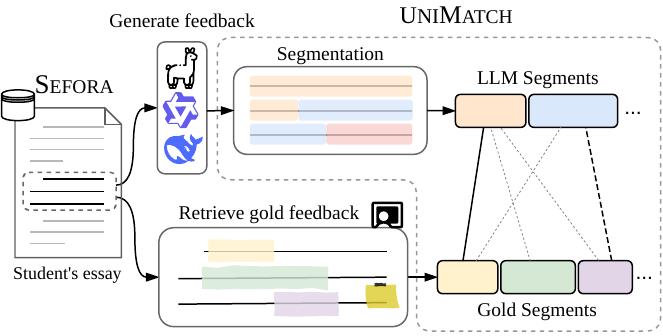}};

\begin{scope}[x={(img.south east)},y={(img.north west)}]
\end{scope}
\end{tikzpicture}
\caption{Overview of the \ourMethod{} evaluation pipeline. For each paragraph, LLM-generated feedback is segmented into units and compared with instructor feedback units. The resulting semantic similarity scores are used to compute an optimal matching between instructor and model feedback units, producing the final evaluation metrics.}\label{fig:overview}
\end{figure}

Progress on this problem is constrained by two bottlenecks. First, few public datasets preserve how instructor feedback is actually delivered in coursework (Table~\ref{tab:dataset-comparison}): multifaceted comments (often addressing several points at once) anchored to specific spans (a paragraph, sentence, or word) and interpretable alongside the assignment prompt, rubric, and revision history. Some resources substitute structured labels (error tags or analytic scores) for formative commentary \cite{crossley2024large, mathias2018asap++, dahlmeier2013building, lee2015cityu}, restrict coverage to a single prompt or narrow genre \cite{kashefi2022argrewrite, zyska2026expos}, or forgo expert annotation for crowd- or model-generated feedback \cite{behzad2024leaf}. Without datasets that capture feedback as instructors deliver it, evaluating whether LLMs produce what an instructor would write remains out of reach.

Second, feedback evaluation is difficult for three reasons. (i) Reference-free evaluation is conceptually appealing, but `good' writing feedback has no single operational definition, as its quality depends on many interacting dimensions weighted differently across raters and settings \cite{pearson2022typology, stahl2024exploring};  LLM-as-judge proxies inherit this ambiguity while adding biases of their own \cite{zheng2023judging, deutsch2022limitations}. (ii) Reference-based evaluation mitigates this by comparing generated feedback against instructor feedback, but existing metrics fall short: n-gram overlap is unreliable for open-ended generation \cite{novikova2017we, reiter2018structured}, and embedding-based metrics, while less surface-bound, correlate only weakly with human judgments \cite{liu2016not}, as we also find in our setting (\S\ref{sec:feedbackSimilarity}). (iii) A deeper, often overlooked issue is granularity: a feedback message typically bundles several distinct \emph{feedback units} \cite{yen2020decipher, zou2024investigating} -- self-contained statements about specific aspects of the writing -- where even a single sentence may contain several (e.g., `This paragraph is well-written, but try to make it more concise'). Holistic comparison cannot tell which observations are recovered or missed, so reliable evaluation must compare feedback at the unit level and by its underlying meaning.

We address both. \ourData{} (\textbf{S}tudent \textbf{E}ssays with \textbf{F}eedback C\textbf{or}pus) is a public corpus pairing authentic instructor-authored span-anchored feedback with assignment prompts, rubrics, analytic scores, and multi-draft revisions across diverse college writing genres. \ourMethod{} is a reference-based framework that evaluates LLM feedback at the unit level rather than holistically, segmenting feedback into feedback units and aligning them via optimal matching under instructor-derived similarity criteria; the resulting similarity-weighted F$_1$ captures both what the model omits and what it over-generates (Figure~\ref{fig:overview}). Together they enable systematic study of LLM-generated writing feedback at the granularity instructors work in.

Our experiments across 74 configurations spanning multiple LLMs show that no setting exceeds $0.4$ F$_1$. Models struggle to produce the comments instructors would emphasize. Feedback verbosity, in the sense of producing more comments rather than longer ones, is the dominant driver of poor alignment. What \ourMethod{} penalizes is pedagogically meaningful: overly verbose feedback overwhelms the student, leading to counterproductive results~\cite{kluger1996effects}.

\section{Related Work}\label{sec:related_work}
\newcommand{\cmark}{\ding{51}}
\newcommand{\xmark}{\ding{55}}
\begin{table*}
  \centering
  \small
  \setlength{\tabcolsep}{6pt}
  \renewcommand{\arraystretch}{1.15}
  \begin{tabularx}{\textwidth}{l c c c c X}
  \toprule
    \multirowcell{2}[0pt][l]{\textbf{\makecell{\\Dataset name}}}
      & \multirowcell{2}{\textbf{\makecell{\\Size\\(\# drafts)}}}
      & \multicolumn{3}{c}{\textbf{Feedback word count}}
      & \multirowcell{2}[0pt][l]{\textbf{\makecell{\\Writing genre(s)}}} \\
    \cmidrule(lr){3-5}
    & & \textbf{\makecell{Total\\(corpus)}} & \textbf{\makecell{Inline\\/draft}} & \textbf{\makecell{Overall\\/draft}} & \\
    \midrule
    Expos\'ia\ \scriptsize{\citep{zyska2026expos}}          & 55  & 17.6K  & 107 & 213 & Proposal \\
    ArgRewrite V.2\ \scriptsize{\citep{kashefi2022argrewrite}}    & 86  & 12.9K   & --  & 150  & Essay \\
    Insta-Reviewer\ \scriptsize{\citep{jia2022insta}}    & 484 & 26.6K  & --  & 55  & Research Report \\
    \textbf{\ourData{}} \scriptsize{(ours)} & 564 & 136.4K & 147 & 95
       & Essay, Narrative, Explanation, Empathy \\
    \bottomrule
  \end{tabularx}
  \caption{\label{tab:dataset-comparison} Comparison of related datasets of student writing with instructor feedback. The table counts only instructor-authored feedback (peer, user, and AI-augmented excluded), with per-draft word counts averaged over each corpus's total draft count. Writing genres follow the BAWE classification of \citet{gardner2013classification}. Inline denotes span-anchored feedback; other features (e.g., error tags, scores) are omitted.}
  \end{table*}

\paragraph{Essay datasets.}\label{related:dataset}
Many existing writing datasets target \emph{Automatic Essay Scoring} (AES), predicting numerical scores rather than formative feedback \citep{ramesh2022automated, ke2019automated, hou2025improve}, or emphasize structured annotations (e.g., error tags or analytic scores) over formative text, including PERSUADE 2.0 \citep{crossley2024large}, ASAP/ASAP++\footnote{\url{https://www.kaggle.com/c/asap-aes}} \citep{mathias2018asap++}, and NUCLE \citep{dahlmeier2013building}. While valuable for large-scale assessment, scores rate a draft but do not tell a student how to revise it \citep{ke2019give}; feedback generation instead requires localized, actionable, context-sensitive comments tied to the writer's text and the assignment goals.

Efforts to collect feedback-based essay datasets have grown in recent years, yet each captures only part of the setting we target. They pair expert feedback with a narrow slice of writing: Insta-Reviewer \citep{jia2022insta} with graduate project reports, ArgRewrite V.2 \citep{kashefi2022argrewrite} with a single argumentative prompt, and Expos\'ia \citep{zyska2026expos} with research expos\'es, while CityU \citep{lee2015cityu}\footnote{The original paper described the corpus as not yet publicly available at the time, and we could not verify a current public release.} offers short tutor comments on ESL/EFL writing. Each is limited on at least one axis central to classroom feedback, lacking multiple drafts or revisions \citep{jia2022insta, lee2015cityu} or covering only a single genre \citep{jia2022insta, kashefi2022argrewrite, zyska2026expos}. LEAF \citep{behzad2024leaf} scales to more essay-feedback pairs but sources them from online users and AI rather than instructors. \ourData{} complements these resources with authentic, instructor-authored feedback, both span-anchored and overall, at larger scale across diverse college writing genres (Table~\ref{tab:dataset-comparison}), paired with assignment prompts, rubrics, analytic scores, and a multi-draft revision structure, enabling analyses that prior corpora do not directly support.

\paragraph{Feedback evaluation.}\label{related:evaluation}
Evaluating generated feedback is difficult, as wording and delivery can vary widely even for the same underlying issue. Reference-free evaluation raises the issues discussed previously: effective feedback has no single operational definition \cite{pearson2022typology, stahl2024exploring}, and LLM-as-judge proxies inherit this ambiguity while adding biases of their own \cite{zheng2023judging, deutsch2022limitations}. We thus focus on reference-based and human-centered evaluation, which fall into three groups \cite{celikyilmaz2020evaluation}. Content-overlap metrics such as BLEU \cite{papineni2002bleu} and ROUGE \cite{lin2004rouge} reward surface similarity and are brittle for open-ended feedback. Model-based metrics such as BERTScore \cite{zhang2019bertscore}, BARTScore \cite{yuan2021bartscore}, and BLEURT \cite{sellam2020bleurt} capture general semantics better but are hard to interpret and miss whether LLM feedback identifies the same substantive points as the instructor. Human-centered evaluation is reliable but costly and hard to scale \cite{jia2022insta}. \ourMethod{} targets both reliability and interpretability by comparing feedback as individual units and scoring their correspondence under feedback-specific similarity criteria.

\section{Dataset}\label{sec:dataset}

\begin{figure*}[t]
\centering
\begin{tikzpicture}
\node[anchor=south west, inner sep=0] (img) at (0,0)
    {\includegraphics[width=\linewidth]{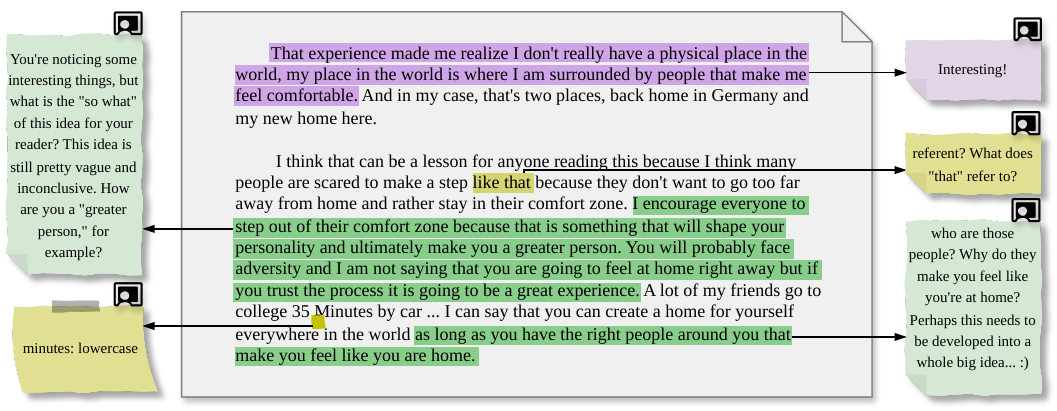}};

\begin{scope}[x={(img.south east)},y={(img.north west)}]
\end{scope}
\end{tikzpicture}
\caption{An annotated draft from \ourData{}, illustrating the dataset's span-anchored, fine-grained instructor feedback: sticky-note comments paired with color-coded highlights. Pink highlights mark especially effective passages, yellow highlights indicate issues needing attention, and green highlights mark ideas worth developing in a subsequent draft. Full annotation conventions are documented in \S\ref{sec:annotation_details}.}
\label{fig:dataSample}
\end{figure*}

\ourData{} was collected over two semesters from a range of first-year undergraduate courses in the university's English Department, spanning composition, ESL, narrative and creative writing, and a senior seminar. It includes assignment prompts, grading rubrics, multi-draft student essays, and two forms of instructor feedback: span-anchored (inline) comments (Figure~\ref{fig:dataSample}) and overall assessments. The dataset is publicly available.\footnote{\ourDataLink{}}

\paragraph{Scope, scale, and genres.}
\ourData{} contains 371 essays by 155 students across 34 assignments from 9 classes, comprising 564 drafts (220 single-draft, 109 two-stage, 42 three-stage) that total 8{,}186 paragraphs. Under the BAWE classification \citep{gardner2013classification}, these span four genre families: Essay, Narrative Recount, Explanation, and Empathy Writing.\footnote{For each assignment we map its writing prompt to the genre family it most closely matches; BAWE's explicit definitions and examples make this a direct mapping.} Instructors contributed 5{,}684 inline (span-anchored) annotations -- highlights, sticky notes, and strikeouts, nearly all with comment -- averaging 10 annotations and 147 feedback words per draft. Additionally, 437 drafts carry overall comments (121 words avg.), 295 carry analytic trait scores (6.5 per draft avg.), and 205 a holistic score, for 8{,}240 instructor annotations in total.

\paragraph{Annotation content.}
\ourData{} preserves the original paragraph structure and aligns each annotation to the text it targets: most are paragraph-level and \emph{span-anchored} -- tied to the exact span the instructor marked, whether a phrase, sentence, or word -- with document-level comments and scores retained when present (details in \S\ref{sec:annotation_details}). What distinguishes \ourData{} from error-tag and score-only corpora is the \emph{substance} of its feedback: it is overwhelmingly \emph{facilitative} (inviting the writer's own reflection) rather than \emph{directive} (prescribing changes) \cite{straub1996concept}. Rather than flagging surface errors, instructors engage substantive aspects of writing -- narrative development, idea elaboration, reader engagement, voice, revision guidance -- and frequently pose open questions back to the writer.

\paragraph{Collection, Privacy, and Ethics.}\label{sec:ethics}
Data collection was approved by the university's IRB. The research team contacted instructors, who emailed students a recruitment notice outlining the study, its objectives, risks, and voluntary participation. Materials were gathered from instructors and consenting students across two semesters. All IRB-related materials are provided in \S\ref{sec:data_collection}.
All essays were released with student consent and processed under strict privacy constraints. We remove direct identifiers (student and instructor names, course identifiers, submission dates) and replace names within essay bodies with consistent within-essay pseudonyms while preserving grammatical and narrative coherence; references to public figures are retained. We further screen the corpus for indirect identifying information (e.g., social media or sandbox handles), and check for abusive or harmful content using \texttt{omni-moderation}\footnote{\url{https://developers.openai.com/api/docs/models/omni-moderation-latest}} exclusively via the API under OpenAI's data-handling policy.\footnote{\url{https://developers.openai.com/api/docs/guides/your-data/}} We do not mask vulgar or strong language, as doing so would alter the writer's voice and distort the feedback context; the dataset is released with appropriate content warnings.

\paragraph{Parsing.}
Original submissions in PDF or \texttt{.docx} were converted to a JSON format that preserves both paragraph segmentation and annotation anchoring. We release the deterministic parser\footnote{\ourMethodLink{}} to support future work on similar materials. Full parsing details are in \S\ref{sec:parsing_details}.

\section{Evaluation Framework}\label{sec:eval}
Beyond a dataset, scaling instructor-quality feedback requires a way to evaluate it; both reference-free and reference-based approaches face challenges. We introduce \ourMethod{}, a reference-based framework that compares model and instructor feedback at the level of feedback units rather than as monolithic texts. Instructor feedback serves as a principled reference for two reasons. First, it encodes pedagogical priorities -- what to flag, what to praise, and how to phrase guidance -- per curriculum rubrics and teaching goals. Second, good writing feedback has no single operational definition~\cite{pearson2022typology, stahl2024exploring}; instructor judgment, while not the only valid target, is the most defensible anchor for a given draft, especially for the open-ended writing in \ourData{} where LLM feedback remains far from human quality \citep{chakrabarty2024art, gomez2023confederacy}.

\ourMethod{} compares model and instructor feedback in three stages (Figure~\ref{fig:overview}): (i) segment each into feedback units, (ii) score semantic correspondence between instructor-model unit pairs, and (iii) apply maximum bipartite matching over those scores to derive interpretable unit-level alignments and aggregate precision, recall, and F$_1$. Stages (i) and (ii) are independently verifiable against human annotation, enabling robustness checks on the pipeline.

\subsection{Feedback Segmentation}\label{sec:segmentation}
Prior work has typically treated feedback as a monolithic message rather than a set of separable units \cite{wu2020feedback,lyu2024steps}, but holistic comparison cannot tell which individual observations are recovered or missed. Other work has recognized the importance of feedback units, though with varying definitions: one or more sentences expressing a coherent thought \citep{yen2020decipher}, or a self-contained message targeting a specific issue \citep{zou2024investigating}. Building on these \citep{yen2020decipher, zou2024investigating}, we adopt \emph{feedback unit} -- a self-contained statement addressing one specific aspect of the student's writing -- as the basic evaluation granularity, and develop an annotation guideline (\S\ref{sec:segmentation_guidelines}) for segmenting instructor and LLM-generated feedback (examples in Table~\ref{tab:segmentationExamples}).

This task is related to discourse parsing \cite{li2014recursive,soricut2003sentence,marcu2000theory}, but differs in two key ways: we omit inter-segment relations, and place emphasis on sub-sentence segmentation, as a single sentence may contain multiple feedback units, while a single unit may span multiple sentences. We therefore draw on prior work on sub-sentence segmentation in discourse \cite{li2022survey,stab2017parsing,hua2019argument,wang2018toward}.

\paragraph{Formalization and agreement metrics.}
Let $T=t_1,\dots,t_n$ be the sequence of words of a feedback message. A segmentation is represented by a boundary vector $Q\in\{0,1\}^{n-1}$ where $Q_i=1$ marks a boundary between $t_i$ and $t_{i+1}$. We measure segmentation agreement using WindowDiff \cite{pevzner2002critique}, a standard discourse segmentation metric, with window size $k$ set to half the average reference segment length ($k{=}10$ in our samples):
\[
WD(r,h)=\frac{1}{n-k}\sum_{i=1}^{n-k}
\mathbf{1}\!\left[b(r_{i:i+k}) \neq b(h_{i:i+k})\right],
\]
where $r$ and $h$ are the two segmentations being compared, one designated the reference and the other the hypothesis (the second annotator, or the model when validating automation), and $b(x_{i:j})$ is the boundary count between $t_i$ and $t_j$ in segmentation $x$. Because WindowDiff tolerates small boundary shifts, we also report boundary precision, recall, and F$_1$ calculated over boundary indices while excluding true negatives to avoid inflated agreement.

\paragraph{Automatic segmentation validation.}
Following the segmentation guideline (\S\ref{sec:segmentation_guidelines}), two graduate students independently annotated more than 50 feedback instances, containing both instructor and LLM feedback. Agreement was high: WD =$0.073$, precision =$0.99$, recall =$0.94$, F$_1$=$0.96$. The annotators then resolved disagreements to produce a consensus segmentation, which we treat as the reference.

Using the same segmentation guideline given to human annotators as prompt, GPT-5-nano reaches comparable agreement against this consensus reference (WD =$0.036$, precision =$1.00$, recall =$0.95$, F$_1$=$0.97$), indistinguishable from inter-annotator agreement.

\newlength{\fuBoxH}
\setlength{\fuBoxH}{2.6ex} 

\newlength{\fuStepY}
\setlength{\fuStepY}{16pt} 

\tikzset{
  fuBoxStyle/.style={
    draw,
    rounded corners=1pt,
    inner xsep=1pt,
    inner ysep=0pt,          
    minimum height=\fuBoxH,  
    text height=1.6ex,       
    text depth=0.4ex         
  },
  fuLineStyle/.style={
    draw=none,
    inner xsep=1pt,
    inner ysep=0pt,
    minimum height=\fuBoxH,
    text height=1.6ex,
    text depth=0.4ex
  }
}

\newcommand{\fuBox}[1]{%
  \tikz[baseline=(n.base)]\node[fuBoxStyle] (n) {#1};%
}

\begin{table}[t]
\centering
\small
\setlength{\tabcolsep}{6pt}
\begin{tabular}{@{} >{\raggedright\arraybackslash}p{0.97\columnwidth}@{}}
\toprule
\textbf{\# \ \ Segmented feedback} \\
\midrule

\textbf{1: }\fuBox{This is well-written,} \fuBox{but try to make it more concise.} \\
\vspace{-1.5pt}

\textbf{2: }\fuBox{Nice! So much authenticity in your voice.} \\
\vspace{-1.5pt}

\textbf{3: }\tikz[baseline=(t.base)]{
  \node[fuLineStyle, anchor=base west] (t) at (0,0)
    {Try to add more sensory details. Like what was special};
  \node[fuLineStyle, anchor=base west] (b) at (0,-\fuStepY-0pt)
    {about that night?};
\draw[rounded corners=1pt]
  (t.north west) -- (t.north east) -- ($(t.south east)+(0pt,2pt)$)
  -- ($(t.south east |- b.north)+(0pt,2pt)$)
  -- ($(b.north east)+(0pt,2pt)$) -- ($(b.south east)+(0pt,0.4pt)$) -- ($(b.south west)+(0pt,0.4pt)$)
  -- (b.north west) -- (t.south west) -- cycle;
  \node[fuBoxStyle, anchor=base west] (also) at ([xshift=1.5pt, yshift=-2.3pt]b.east)
    {Also, why did he suddenly leave?};
} \\

\bottomrule

\end{tabular}
\caption{Examples of feedback segmented into feedback units.}
\label{tab:segmentationExamples}
\end{table}

\subsection{Feedback Unit Similarity}\label{sec:feedbackSimilarity}
We then compare the similarity between feedback units. This task is related to Semantic Textual Similarity (STS) \cite{bar2012ukp,majumder2016semantic}, but differs in two key ways: the texts being compared are feedback units rather than arbitrary sentences, and both address the same paragraph of the same essay.

We thus introduce a task-specific guideline (\S\ref{sec:similarity_guidelines}) for scoring feedback unit similarity along two dimensions of feedback: the \emph{target} (the aspect, span, or idea addressed) and the \emph{comment} (the evaluation, critique, or suggestion about that target). Drawing on integer-scale scoring practice in STS \cite{agirre2012semeval,xu2015semeval,agirre2015semeval}, we use a 0--4 scale chosen to suit the feedback setting: anchored at 4 (near-equivalent as feedback) and 0 (irrelevant or contradictory), with an explicit middle value (2) for pairs that share an important point but differ on another (intermediate levels and details in \S\ref{sec:similarity_guidelines}). The guideline was developed in collaboration with course instructors and refined iteratively by two graduate student annotators, who piloted it, discussed disagreements, and refined wording and examples until agreement stabilized. They then independently scored 120 feedback unit pairs spanning the full scale; we report inter-annotator Pearson and Spearman correlations of $r=0.7572$ and $\rho=0.7558$. We use the mean as the reference similarity for these 120 pairs.

\begin{figure}[t]
\centering
\begin{tikzpicture}[scale=0.7]

\pgfplotstableread[col sep=space]{figures/heatmap_matrix.dat}\datatable

\foreach \r in {0,1,2,3,4}{%
  \foreach \c in {0,1,2,3,4}{%
    \pgfplotstablegetelem{\r}{[index]\c}\of\datatable
    \pgfmathsetmacro{\val}{\pgfplotsretval}

    \pgfmathsetmacro{\shade}{100*\val/24}

    \fill[black!\shade] (\c,4-\r) rectangle ++(1,1);

    \draw[thin] (\c,4-\r) rectangle ++(1,1);
  }%
}%

\foreach \i in {0,1,2,3,4}{%
  \pgfmathtruncatemacro{\gt}{4-\i}
  \node[anchor=east] at (0.00,4.5-\i) {\small{annot=\gt}};
}

\foreach \c in {0,1,2,3,4}{%
  \node[anchor=north, rotate=60] at (\c+0.13,-0.6) {\small{pred=\c}};
}
\end{tikzpicture}
\caption{Agreement between annotator-averaged scores (rows, rounded to the model's scale) and \texttt{gemini-3.1-flash-lite-preview} scores (columns); darker cells hold more pairs. Intensity along the diagonal indicates strong agreement; the low-score corner is darkest as most feedback-unit pairs are unrelated and score low.}\label{fig:heatmap}
\end{figure}
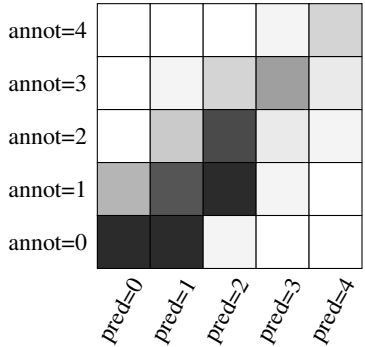

\paragraph{Automatic similarity scorer validation.} Existing similarity methods fall short on this task: lexical-overlap metrics yield Pearson correlations below $0.1$ and embedding-based methods reach at most $r\approx 0.67$, both too low to be reliable (details in \S\ref{sec:automatic_similarity_metrics_appendix}). We therefore experimented with several LLMs using the defined guideline as prompt, with several exceeding $r=0.7$ and the strongest exceeding $r=0.8$. However, as scoring every pair independently at our scale would be expensive, we introduce batching (details in \S\ref{sec:batch_details}), which preserves performance at a fraction of the cost (Table~\ref{tab:similarityAgreement}\footnote{Model performance can shift over time; results were collected in May 2026. Result using the snapshot model \texttt{gpt-5-mini-2025-08-07} is also reported for stability.}). We adopt \texttt{gemini-3.1-flash-lite-preview} as the similarity scorer in our main pipeline ($r=0.813$ and $\rho=0.818$ against the human reference). Figure~\ref{fig:heatmap} visualizes its agreement on a sample with the pipeline's natural score distribution.

\begin{table}
\small
\centering
\begin{tabular}{@{}llllcc@{}}
\toprule
\textbf{Model} & \textbf{Version} & \textbf{Size} & \textbf{Reason} & $\bm{r}$ & $\bm{\rho}$ \\
\midrule
\multirow{3}{*}{GPT}
  & 5.1               & --         & low     & $0.805$ & $0.794$ \\
  & 5$^{\dagger}$     & mini       & low     & $0.723$ & $0.730$ \\
  & 5                 & nano       & low     & $0.598$ & $0.604$ \\
\midrule
\multirow{2}{*}{Gemini}
  & 3.1               & flash-lite$^{\ddagger}$ & minimal & $0.813$ & $0.818$ \\
  & 2.5               & lite       & --    & $0.701$ & $0.711$ \\
\bottomrule
\end{tabular}
\caption{The Pearson $r$ and Spearman $\rho$ correlations between LLM-generated and human-annotated similarity scores for paired feedback units (with \texttt{batch\_size}$=50$) over $120$ samples. $^{\dagger}$Snapshot \texttt{gpt-5-mini-2025-08-07}. $^{\ddagger}$ Model name \texttt{gemini-3.1-\allowbreak flash-lite-preview}.}
\label{tab:similarityAgreement}
\end{table}

\subsection{Feedback Unit Cross Matching}\label{sec:crossmatching}
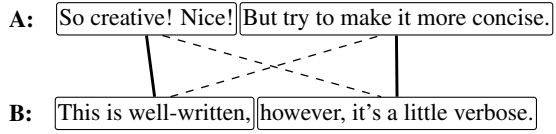
\begin{figure}[t]
\small

\newlength{\rowy}
\setlength{\rowy}{1.25cm}

\newlength{\boxh}
\setlength{\boxh}{3ex}

\newlength{\labelgap}
\setlength{\labelgap}{5pt}

\begin{tikzpicture}[
  box/.style={
    draw,
    rounded corners=1pt,
    inner xsep=1pt,
    inner ysep=0pt,
    minimum height=\boxh,
    text height=1.6ex,
    text depth=0.4ex,
    align=left,
    font=\small
  }
]

\node[anchor=west] (labelA) at (0,0) {\textbf{A:}};
\node[anchor=west] (labelB) at (0,-\rowy) {\textbf{B:}};

\path (labelA.east); \pgfgetlastxy{\xa}{\ya}
\node[box, anchor=west] (nice) at (\xa+\labelgap,\ya) {So creative! Nice!};

\path (nice.east); \pgfgetlastxy{\xe}{\ye}
\node[box, anchor=west] (concise) at (\xe+1pt,\ye)
{But try to make it more concise.};

\path (labelB.east); \pgfgetlastxy{\xb}{\yb}
\node[box, anchor=west] (creative) at (\xb+\labelgap,\yb)
{This is well-written,};

\path (creative.east); \pgfgetlastxy{\x2}{\y2}
\node[box, anchor=west] (verbose) at (\x2+1pt,\y2)
{however, it's a little verbose.};

\draw[line width=1.1pt] (nice.south) -- (creative.north);
\draw[line width=1.1pt] (concise.south) -- (verbose.north);

\draw[dashed, line width=0.5pt] ($(nice.south)+(6pt,0)$) -- ($(verbose.north)+(-6pt,0)$);
\draw[dashed, line width=0.5pt] ($(creative.north)+(6pt, 0)$) -- ($(concise.south)+(-6pt, 0)$);

\end{tikzpicture}

\caption{Matching feedback \textbf{A} and \textbf{B}: boxes mark segments; solid lines = high similarity, dashed = low. }
\label{fig:segMatch}
\end{figure}

Given pairwise similarity scores between instructor and model-generated feedback units, \ourMethod{} computes a one-to-one alignment between the two sets. Intuitively, each feedback unit is matched to its most similar counterpart on the other side, subject to a globally optimal overall matching (Figure~\ref{fig:segMatch} illustrates). Specifically, for instructor feedback units $G=\{G_i\}_{i=1}^{|G|}$ and predicted feedback units $P=\{P_j\}_{j=1}^{|P|}$, let $\phi(G_i, P_j) \ge 0$ denote the similarity scorer. We define the optimal alignment as the maximum-weight bipartite matching
\[
M^*=\arg\max_{M\in\mathcal{M}} \sum_{(G_i,P_j)\in M}\phi(G_i,P_j),
\]
where $\mathcal{M}$ is the set of 1-to-1 matchings between $G$ and $P$. Computing $M^*$ is a standard maximum-weight bipartite matching problem, and we use the Hungarian algorithm \cite{kuhn1955hungarian} for efficiency. Let $\Phi(M^*)=\sum_{(G_i,P_j)\in M^*}\phi(G_i,P_j)$ denote the total weight of this alignment.

To turn $\Phi(M^*)$ into interpretable precision and recall, we normalize it against per-side upper bounds, each set's summed within-set self-similarity ($\sum_j \phi(P_j,P_j)$ predicted, $\sum_i \phi(G_i,G_i)$ instructor), which equals its unit count times the maximum score (4) since each unit matches itself. Following coreference resolution~\cite{luo2005coreference}, we define soft precision and recall as
\[
p=\frac{\Phi(M^*)}{\sum_j \phi(P_j,P_j)}, \qquad
r=\frac{\Phi(M^*)}{\sum_i \phi(G_i,G_i)},
\]
and the soft F$_1$ as
\[
F_1=\frac{2pr}{p+r}.
\]
The result is an interpretable, soft precision/recall view of feedback quality: precision captures how much of the model's feedback aligns with what an instructor wrote, and recall captures how much of the instructor's feedback the model recovers.\footnote{The pipeline also admits a threshold-based instantiation with hard matches above a chosen $\tau$, exposing strictness as an explicit knob (\S\ref{sec:appendix_threshold}).}

\paragraph{End-to-end validation.}
\ourMethod{}'s components -- segmentation (\S\ref{sec:segmentation}) and similarity scoring (\S\ref{sec:feedbackSimilarity}) -- are validated against human judgment in isolation, but this does not guarantee that the composed metric aligns with instructor judgment. To test the pipeline end to end, we selected 20 paragraphs, each with three raw LLM feedback texts (\S\ref{sec:experiments}) that \ourMethod{} scores at well-separated levels (F$_1\approx 0$, $0.5$, $1$), and had a writing expert independently rank the three by quality against the reference feedback.\footnote{Ties were permitted but occurred for only two paragraphs.} Across all 60 texts, \ourMethod{} F$_1$ correlates strongly with the instructor ranking (Pearson $r=0.836$, Spearman $\rho=0.837$), showing that the full pipeline recovers expert judgments.

\section{Experiments}\label{sec:experiments} 
We evaluate paragraph-level LLM-generated feedback under \ourMethod{}, varying model choice and prompting setup to study what task framing and generation choices drive feedback quality.

\paragraph{Task framing and evaluation data.} 
We generate feedback at the paragraph level, matching the granularity of inline instructor annotations in \ourData{}. For each target paragraph, the model receives the paragraph plus as much surrounding context as fits within a fixed 8{,}192-token budget, enough to include the entire essay for most drafts. We evaluate on the 3{,}499 paragraphs in \ourData{} with at least one instructor annotation (\emph{full-corpus}); of these, 2{,}324 carry exactly one (the \emph{single-reference} subset). We report metrics on both.

\paragraph{Models.}
To support broad ablations across settings, we primarily use small open-weight instruction-tuned models, namely, Llama-3.1-8B \cite{grattafiori2024llama} (Llama), DeepSeek-R1-Distill-Llama-8B \cite{guo2025deepseek} (DeepSeek), and Mistral-7B \cite{jiang2023mistral7b} (Mistral). To probe whether the effects extend to a larger regime, we also evaluate Qwen2.5-72B-Instruct \cite{qwen2025qwen25} (Qwen) and closed model GPT-5.1\footnote{\url{https://developers.openai.com/api/docs/models/gpt-5.1}} (GPT) on a subset of configurations.

\paragraph{Prompt template.}
We use two prompt templates. The default \emph{constrained} prompt asks the model to identify the single most important focus in the target paragraph and produce one feedback unit under a prescribed output format; the \emph{unconstrained} prompt places no limit on the number of feedback units it may return. On top of the template, we vary four further dimensions of task specification: zero-shot vs.\ few-shot prompting (5 exemplars) \cite{brown2020language}; guided vs.\ unguided prompting, where the guided setting supplies a list of feedback categories~\cite{narciss2008feedback,keuning2018systematic} in the prompt; inclusion vs.\ omission of the assignment prompt and rubric; and chain-of-thought prompting \cite{wei2022chain,lyu2023faithful}. The guided categories cover the common functional roles of writing feedback: \emph{Task Constraints}, \emph{Concepts}, \emph{Elaboration}, \emph{Clarification}, \emph{Mistakes}, and \emph{Praise}. Few-shot exemplars are 5 instructor-authored feedback samples covering each category. Full prompt templates and guided variants are in \S\ref{app:prompt_templates}.

\paragraph{Generation settings.}
We allocate up to 250 output tokens per target paragraph -- well above the typical length of instructor feedback on a single paragraph in \ourData{} ($\approx20$ tokens\footnote{Exact number of tokens depends on the tokenizer.}) -- to avoid truncating multi-point responses.

In total, the model and prompt combination yields 74 experimental configurations. All prompts (\S\ref{app:prompt_templates}), inference pipelines, the \ourMethod{} framework, and code are publicly available.\footnote{\ourMethodLink{}} %

\section{Results}\label{sec:results}

\definecolor{cmapHigh}{HTML}{2166AC}
\definecolor{cmapLow}{HTML}{B2182B}
\definecolor{cmapHigh}{HTML}{00360E}
\definecolor{cmapLow}{HTML}{FFFFFF}
\definecolor{okabeBlue}{HTML}{0072B2}
\definecolor{okabeVerm}{HTML}{D55E00}
\definecolor{okabeGreen}{HTML}{009E73}

\begin{table*}[]
\small
\centering
\begin{tabular}{@{}lcccccccc@{}}
\toprule
\multirow{2}{*}{\textbf{\vspace{-5pt}Model}} &
  \multirow{2}{*}{\textbf{\vspace{-5pt}Precision}} &
  \multirow{2}{*}{\textbf{\vspace{-5pt}Recall}} &
  \multirow{2}{*}{\textbf{\vspace{-5pt}F$\mathbf{_1}$}} &
  \multicolumn{5}{c}{\textbf{Ablations (F$\mathbf{_1}$)}} \\ \cmidrule(l){5-9} 
 &
   &
   &
   &
  \textbf{Single-reference} &
  \textbf{Unconstrained} &
  \textbf{Unguided} &
  \textbf{No rubrics} &
  \textbf{Few-shot} \\ \midrule
Llama\_8B   & 0.275 &    0.236  & \multicolumn{1}{l||}{0.254} & 0.290 &  0.113 &  0.203 & 0.268 &  0.271 \\
Mistral\_7B & 0.288 &   0.267   & \multicolumn{1}{l||}{0.277} & 0.311 &  0.159 &  0.251 & 0.278 &  0.258 \\
Qwen\_72B   & 0.374 &   0.268   & \multicolumn{1}{l||}{0.312} & 0.387 &  0.179 &  0.271 & 0.285 &  0.317 \\
GPT-5.1      & 0.376 &   0.273   & \multicolumn{1}{l||}{0.316} & 0.371 &  0.136 &  0.296 & 0.288 &  0.307 \\ \bottomrule
\end{tabular}
\caption{\ourMethod{} results for the configuration <zero-shot, rubric, guided, default prompt> on the full corpus. Left: precision, recall, F$_1$. Right: F$_1$ under single-axis variations -- \emph{Single-reference} subset (paragraphs with exactly one instructor annotation),  \emph{Unconstrained} prompt, \emph{Unguided}, \emph{No rubrics}, and \emph{Few-shot}. Full per-configuration results in Tables~\ref{tab:fullResultsZero}--\ref{tab:fullResultsCoT} (\S~\ref{app:full_results}).}\label{tab:mainResults} 
\end{table*}

Table~\ref{tab:mainResults} reports \ourMethod{} scores on the full corpus under the zero-shot setting (with rubrics and category guidance), where most models perform best; full results are in Tables~\ref{tab:fullResultsZero}--\ref{tab:fullResultsCoT} of \S\ref{app:full_results}. No configuration exceeds $0.4$ F$_1$, and both precision and recall stay low: even the strongest GPT model reaches only $0.38$ precision and $0.27$ recall, so models miss most of the feedback instructors prioritize while spending much of their output on points instructors do not raise. Among the small models ($\le8$B parameters), Mistral is consistently strongest, ahead of Llama\footnote{Sometimes DeepSeek as shown in \S\ref{app:full_results}.}. The larger Qwen and GPT lead overall -- GPT on the shared subset, Qwen on the single best configuration ($0.32$ F$_1$) -- but only narrowly, and all remain far from matching instructor feedback.

Table~\ref{tab:mainResults} additionally presents ablations: each varies a single variable relative to the base setting <zero-shot, rubric, guided, default prompt>. We show a subset due to space constraints, but the pattern matches the full results. On the single-reference subset (fourth column), scores are higher as expected, since our constrained default prompt emits one feedback unit per paragraph and thus naturally aligns more with single-reference paragraphs.

\paragraph{Verbosity.}

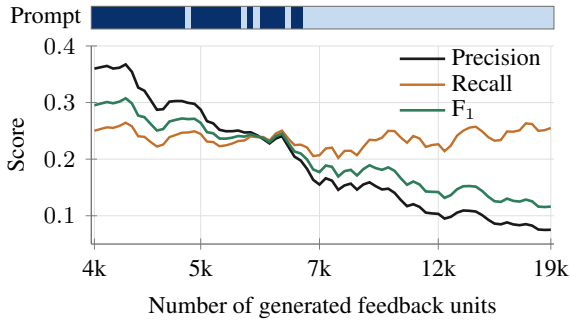
\begin{figure}[t]
\centering
\definecolor{promptV2}{HTML}{08306B}   
\definecolor{promptV1}{HTML}{C6DBEF}   
\definecolor{cPrec}{HTML}{1A1A1A}
\definecolor{cRec}{HTML}{C4702A}
\definecolor{cF1}{HTML}{2E7D5B}
\begin{tikzpicture}
\begin{axis}[
  width=\columnwidth, height=4.2cm,
  xmin=0.5, xmax=74.5,
  xtick={1,18,37,56,74}, xticklabels={4k,5k,7k,12k,19k},
  ymin=0.05, ymax=0.40, ytick={0.1,0.2,0.3,0.4},
  axis x line=bottom, axis y line=left,
  axis line style={draw=black!55, line width=0.5pt, -},   
  xlabel={Number of generated feedback units}, ylabel={Score},
  tick label style={font=\footnotesize}, label style={font=\small},
  grid=major, major grid style={line width=0.2pt, draw=black!12},
  legend style={draw=none, fill=none, font=\small,
                at={(0.99,1.03)}, anchor=north east, row sep=-2pt, inner sep=1pt},
  legend cell align={left},
  clip=false,
]
\addplot[cPrec, line width=1.0pt] table[x=Rank,y=Precision] {figures/prf_smooth.dat};
\addlegendentry{Precision}
\addplot[cRec, line width=0.9pt] table[x=Rank,y=Recall] {figures/prf_smooth.dat};
\addlegendentry{Recall}
\addplot[cF1, line width=1.0pt] table[x=Rank,y=F1] {figures/prf_smooth.dat};
\addlegendentry{F$_1$}

\fill[promptV2] (axis cs:0.5,0.43) rectangle (axis cs:1.5,0.47);
\fill[promptV2] (axis cs:1.5,0.43) rectangle (axis cs:2.5,0.47);
\fill[promptV2] (axis cs:2.5,0.43) rectangle (axis cs:3.5,0.47);
\fill[promptV2] (axis cs:3.5,0.43) rectangle (axis cs:4.5,0.47);
\fill[promptV2] (axis cs:4.5,0.43) rectangle (axis cs:5.5,0.47);
\fill[promptV2] (axis cs:5.5,0.43) rectangle (axis cs:6.5,0.47);
\fill[promptV2] (axis cs:6.5,0.43) rectangle (axis cs:7.5,0.47);
\fill[promptV2] (axis cs:7.5,0.43) rectangle (axis cs:8.5,0.47);
\fill[promptV2] (axis cs:8.5,0.43) rectangle (axis cs:9.5,0.47);
\fill[promptV2] (axis cs:9.5,0.43) rectangle (axis cs:10.5,0.47);
\fill[promptV2] (axis cs:10.5,0.43) rectangle (axis cs:11.5,0.47);
\fill[promptV2] (axis cs:11.5,0.43) rectangle (axis cs:12.5,0.47);
\fill[promptV2] (axis cs:12.5,0.43) rectangle (axis cs:13.5,0.47);
\fill[promptV2] (axis cs:13.5,0.43) rectangle (axis cs:14.5,0.47);
\fill[promptV2] (axis cs:14.5,0.43) rectangle (axis cs:15.5,0.47);
\fill[promptV1] (axis cs:15.5,0.43) rectangle (axis cs:16.5,0.47);
\fill[promptV2] (axis cs:16.5,0.43) rectangle (axis cs:17.5,0.47);
\fill[promptV2] (axis cs:17.5,0.43) rectangle (axis cs:18.5,0.47);
\fill[promptV2] (axis cs:18.5,0.43) rectangle (axis cs:19.5,0.47);
\fill[promptV2] (axis cs:19.5,0.43) rectangle (axis cs:20.5,0.47);
\fill[promptV2] (axis cs:20.5,0.43) rectangle (axis cs:21.5,0.47);
\fill[promptV2] (axis cs:21.5,0.43) rectangle (axis cs:22.5,0.47);
\fill[promptV2] (axis cs:22.5,0.43) rectangle (axis cs:23.5,0.47);
\fill[promptV2] (axis cs:23.5,0.43) rectangle (axis cs:24.5,0.47);
\fill[promptV1] (axis cs:24.5,0.43) rectangle (axis cs:25.5,0.47);
\fill[promptV2] (axis cs:25.5,0.43) rectangle (axis cs:26.5,0.47);
\fill[promptV1] (axis cs:26.5,0.43) rectangle (axis cs:27.5,0.47);
\fill[promptV2] (axis cs:27.5,0.43) rectangle (axis cs:28.5,0.47);
\fill[promptV2] (axis cs:28.5,0.43) rectangle (axis cs:29.5,0.47);
\fill[promptV2] (axis cs:29.5,0.43) rectangle (axis cs:30.5,0.47);
\fill[promptV2] (axis cs:30.5,0.43) rectangle (axis cs:31.5,0.47);
\fill[promptV1] (axis cs:31.5,0.43) rectangle (axis cs:32.5,0.47);
\fill[promptV2] (axis cs:32.5,0.43) rectangle (axis cs:33.5,0.47);
\fill[promptV2] (axis cs:33.5,0.43) rectangle (axis cs:34.5,0.47);
\fill[promptV1] (axis cs:34.5,0.43) rectangle (axis cs:35.5,0.47);
\fill[promptV1] (axis cs:35.5,0.43) rectangle (axis cs:36.5,0.47);
\fill[promptV1] (axis cs:36.5,0.43) rectangle (axis cs:37.5,0.47);
\fill[promptV1] (axis cs:37.5,0.43) rectangle (axis cs:38.5,0.47);
\fill[promptV1] (axis cs:38.5,0.43) rectangle (axis cs:39.5,0.47);
\fill[promptV1] (axis cs:39.5,0.43) rectangle (axis cs:40.5,0.47);
\fill[promptV1] (axis cs:40.5,0.43) rectangle (axis cs:41.5,0.47);
\fill[promptV1] (axis cs:41.5,0.43) rectangle (axis cs:42.5,0.47);
\fill[promptV1] (axis cs:42.5,0.43) rectangle (axis cs:43.5,0.47);
\fill[promptV1] (axis cs:43.5,0.43) rectangle (axis cs:44.5,0.47);
\fill[promptV1] (axis cs:44.5,0.43) rectangle (axis cs:45.5,0.47);
\fill[promptV1] (axis cs:45.5,0.43) rectangle (axis cs:46.5,0.47);
\fill[promptV1] (axis cs:46.5,0.43) rectangle (axis cs:47.5,0.47);
\fill[promptV1] (axis cs:47.5,0.43) rectangle (axis cs:48.5,0.47);
\fill[promptV1] (axis cs:48.5,0.43) rectangle (axis cs:49.5,0.47);
\fill[promptV1] (axis cs:49.5,0.43) rectangle (axis cs:50.5,0.47);
\fill[promptV1] (axis cs:50.5,0.43) rectangle (axis cs:51.5,0.47);
\fill[promptV1] (axis cs:51.5,0.43) rectangle (axis cs:52.5,0.47);
\fill[promptV1] (axis cs:52.5,0.43) rectangle (axis cs:53.5,0.47);
\fill[promptV1] (axis cs:53.5,0.43) rectangle (axis cs:54.5,0.47);
\fill[promptV1] (axis cs:54.5,0.43) rectangle (axis cs:55.5,0.47);
\fill[promptV1] (axis cs:55.5,0.43) rectangle (axis cs:56.5,0.47);
\fill[promptV1] (axis cs:56.5,0.43) rectangle (axis cs:57.5,0.47);
\fill[promptV1] (axis cs:57.5,0.43) rectangle (axis cs:58.5,0.47);
\fill[promptV1] (axis cs:58.5,0.43) rectangle (axis cs:59.5,0.47);
\fill[promptV1] (axis cs:59.5,0.43) rectangle (axis cs:60.5,0.47);
\fill[promptV1] (axis cs:60.5,0.43) rectangle (axis cs:61.5,0.47);
\fill[promptV1] (axis cs:61.5,0.43) rectangle (axis cs:62.5,0.47);
\fill[promptV1] (axis cs:62.5,0.43) rectangle (axis cs:63.5,0.47);
\fill[promptV1] (axis cs:63.5,0.43) rectangle (axis cs:64.5,0.47);
\fill[promptV1] (axis cs:64.5,0.43) rectangle (axis cs:65.5,0.47);
\fill[promptV1] (axis cs:65.5,0.43) rectangle (axis cs:66.5,0.47);
\fill[promptV1] (axis cs:66.5,0.43) rectangle (axis cs:67.5,0.47);
\fill[promptV1] (axis cs:67.5,0.43) rectangle (axis cs:68.5,0.47);
\fill[promptV1] (axis cs:68.5,0.43) rectangle (axis cs:69.5,0.47);
\fill[promptV1] (axis cs:69.5,0.43) rectangle (axis cs:70.5,0.47);
\fill[promptV1] (axis cs:70.5,0.43) rectangle (axis cs:71.5,0.47);
\fill[promptV1] (axis cs:71.5,0.43) rectangle (axis cs:72.5,0.47);
\fill[promptV1] (axis cs:72.5,0.43) rectangle (axis cs:73.5,0.47);
\fill[promptV1] (axis cs:73.5,0.43) rectangle (axis cs:74.5,0.47);
\draw[black!55, line width=0.4pt] (axis cs:0.5,0.43) rectangle (axis cs:74.5,0.47);
\node[anchor=east, font=\footnotesize] at (axis cs:0.5,0.45) {Prompt~};
\end{axis}
\end{tikzpicture}
\caption{Precision, recall, and F$_1$ across the 74 configurations of \S\ref{sec:experiments}, sorted by number of generated feedback units ($x$-axis is ordinal and not to scale). The strip marks the prompt template: the constrained template (dark) and the unconstrained (light). Precision and F$_1$ fall as volume rises while recall does not compensate.}
\label{fig:prf-vs-segments}
\end{figure}

\textbf{The dominant signal in our experiments is verbosity.} Across all models, the constrained prompt (our default) beats the unconstrained alternative, which yields substantially lower F$_1$ (the \emph{unconstrained} column of Table~\ref{tab:mainResults}). Removing the single-feedback constraint lets models emit more than one feedback unit per paragraph, and the resulting over-generation sharply lowers precision. This holds even on the full corpus, where paragraphs often carry multiple instructor units and the unconstrained prompt should in principle have the edge: its extra units add low-similarity matches that depress precision more than they lift recall. Figure~\ref{fig:prf-vs-segments} confirms the pattern across all evaluated configurations -- the number of generated feedback units is strongly negatively correlated with both precision and F$_1$ ($r=-0.850$ and $r=-0.840$, respectively), while the accompanying recall gains are too modest to compensate (full results in \S\ref{app:full_results}).

\paragraph{Guiding the model with feedback categories.}
Supplying explicit feedback categories (the \emph{guided} setting) is the next strongest factor: it improves F$_1$ by about $+0.059$ without rubric and $+0.035$ with rubric on the full-corpus evaluation in matched comparisons (the \emph{Unguided} column of Table~\ref{tab:mainResults}; per-configuration numbers in Tables~\ref{tab:fullResultsZero}--\ref{tab:fullResultsCoT}, and a per-model breakdown in Figure~\ref{fig:effect_of_guidance}, \S\ref{app:full_results}).

\paragraph{Rubric and few-shot prompting.}
Two further levers contribute little (Table~\ref{tab:mainResults}, \emph{No rubrics} and \emph{Few-shot} columns). Rubric inclusion is essentially neutral, with small, unstable differences: in our paragraph-level setting well-framed prompts already supply most of the structure a rubric would contribute, and rubrics largely encode whether a draft is on-task -- which, in authentic coursework, it usually is, leaving little for a rubric to anchor given that \ourData{}'s feedback is facilitative (\S\ref{sec:dataset}), engaging execution rather than rubric conformance. Few-shot prompting is beneficial but secondary, improving F$_1$ by about $+0.017$ on the full corpus, mainly through precision; its gains are small under the default prompt, as exemplars help most when task framing is weak, and once the prompt already induces the desired behavior, demonstrations add little. Larger few-shot gains under the unconstrained alternative are in Table~\ref{tab:fullResultsFew} (\S\ref{app:full_results}); a small chain-of-thought study on the three small models follows the same pattern (\S\ref{app:full_results}). Per-model breakdowns of both effects appear in Figures~\ref{fig:effect_of_rubric} and~\ref{fig:effect_of_fewshot} (\S\ref{app:full_results}).

\section{Conclusion}\label{sec:conclusion}
We introduce \ourData{}, a corpus of authentic instructor feedback on student writing, and \ourMethod{}, a reference-based framework for evaluating generated feedback. Across 74 configurations, no setting exceeds $0.4$ F$_1$: the central challenge for LLMs is not producing feedback but producing the comments an instructor would prioritize. Over-generation is consistently costly, with precision dropping sharply as models generate more while recall rarely compensates, a pattern that is pedagogically meaningful, since overwhelming a student with feedback is itself counterproductive \cite{kluger1996effects}. Validated end-to-end against expert judgment, \ourMethod{} provides a reliable automatic metric for this task, and together with \ourData{} lays a foundation for future work on instructor-aligned feedback generation across diverse college writing genres.

\section{Limitations}
Although \ourMethod{} enables fine-grained analysis, its evaluation operates on feedback units as segmented, without further decomposing compound statements. Some units express multiple claims (e.g., `this paragraph is detailed and smooth') and could in principle be split further. Future work could incorporate automatic claim decomposition to isolate individual feedback propositions at the atomic level.

Our evaluation operates at the paragraph level: the model is given a target paragraph, and \ourMethod{} assesses the feedback it produces rather than whether the model identifies which span warrants comment in the first place. \ourData{}'s span-anchored annotations make span identification directly studiable, but we leave this setting to future work. Our experiments also evaluate against the inline (span-anchored) feedback only, leaving the document-level overall assessments and analytic scores in \ourData{} for future work.

Our alignment uses maximum-weight bipartite matching over all eligible unit pairs. Alternative alignment constraints (e.g., prioritizing globally highest-scoring pairs before enforcing one-to-one matching) may produce different match structures and could be explored as alternative objectives.

Our pipeline relies on closed-source models (GPT-5-nano for segmentation and \texttt{gemini-3.1-flash-lite-preview} for similarity scoring), chosen because they meet the human-agreement bar required for reliable evaluation in our setting. This reliance has practical downsides: hosted models can change or be deprecated over time, and their use introduces cost and reproducibility constraints. A natural next step is to distill or fine-tune smaller open-source models for both the segmentation and similarity stages, using our human-annotated subsets as supervision. Because each stage is independently validated against human annotation (\S\ref{sec:segmentation}, \S\ref{sec:feedbackSimilarity}), open-source replacements can be substituted into the pipeline whenever they reach comparable agreement, without re-validating \ourMethod{} as a whole.

\ourMethod{} treats instructor feedback as ground truth. Reference-free evaluation may also be possible by leveraging methods from the automated essay scoring (AES) literature. For example, generated feedback could be applied to the relevant essay span and the resulting revision evaluated with a quality scorer to estimate feedback effectiveness. Developing such metrics is beyond the scope of this work.

Finally, our study relies primarily on automatic evaluation under the proposed framework and does not include a complementary human evaluation of overall feedback quality. While \ourMethod{} is designed to better reflect semantic, point-level alignment with instructor feedback than whole-text overlap metrics, it does not directly assess dimensions such as helpfulness, pedagogical appropriateness, or actionability as perceived by human readers. A targeted human study, or an LLM-as-judge evaluation calibrated against human judgments, would provide a useful complementary perspective. We leave such validation to future work.

\bibliography{custom}

@article{grattafiori2024llama,
  title={The llama 3 herd of models},
  author={Grattafiori, Aaron and Dubey, Abhimanyu and Jauhri, Abhinav and Pandey, Abhinav and Kadian, Abhishek and Al-Dahle, Ahmad and Letman, Aiesha and Mathur, Akhil and Schelten, Alan and Vaughan, Alex and others},
  journal={arXiv preprint arXiv:2407.21783},
  year={2024}
}

@article{keuning2018systematic,
  title={A systematic literature review of automated feedback generation for programming exercises},
  author={Keuning, Hieke and Jeuring, Johan and Heeren, Bastiaan},
  journal={ACM Transactions on Computing Education (TOCE)},
  volume={19},
  number={1},
  pages={1--43},
  year={2018},
  publisher={ACM New York, NY, USA}
}

@incollection{narciss2008feedback,
  title={Feedback strategies for interactive learning tasks},
  author={Narciss, Susanne},
  booktitle={Handbook of research on educational communications and technology},
  pages={125--143},
  year={2008},
  publisher={Routledge}
}

@article{brown2020language,
  title={Language models are few-shot learners},
  author={Brown, Tom and Mann, Benjamin and Ryder, Nick and Subbiah, Melanie and Kaplan, Jared D and Dhariwal, Prafulla and Neelakantan, Arvind and Shyam, Pranav and Sastry, Girish and Askell, Amanda and others},
  journal={Advances in neural information processing systems},
  volume={33},
  pages={1877--1901},
  year={2020}
}

@inproceedings{agirre2015semeval,
  title={Semeval-2015 task 2: Semantic textual similarity, english, spanish and pilot on interpretability},
  author={Agirre, Eneko and Banea, Carmen and Cardie, Claire and Cer, Daniel and Diab, Mona and Gonzalez-Agirre, Aitor and Guo, Weiwei and Lopez-Gazpio, Inigo and Maritxalar, Montse and Mihalcea, Rada and others},
  booktitle={Proceedings of the 9th international workshop on semantic evaluation (SemEval 2015)},
  pages={252--263},
  year={2015}
}

@inproceedings{xu2015semeval,
  title={Semeval-2015 task 1: Paraphrase and semantic similarity in twitter (pit)},
  author={Xu, Wei and Callison-Burch, Chris and Dolan, William B},
  booktitle={Proceedings of the 9th international workshop on semantic evaluation (SemEval 2015)},
  pages={1--11},
  year={2015}
}

@article{pearson2022typology,
  title={A typology of the characteristics of teachers’ written feedback comments on second language writing},
  author={Pearson, William S},
  journal={Cogent Education},
  volume={9},
  number={1},
  pages={2024937},
  year={2022},
  publisher={Taylor \& Francis}
}

@inproceedings{stahl2024exploring,
  title={Exploring LLM prompting strategies for joint essay scoring and feedback generation},
  author={Stahl, Maja and Biermann, Leon and Nehring, Andreas and Wachsmuth, Henning},
  booktitle={Proceedings of the 19th workshop on innovative use of NLP for building educational applications (BEA 2024)},
  pages={283--298},
  year={2024}
}

@inproceedings{zheng2023judging,
  title     = {Judging {LLM-as-a-Judge} with {MT-Bench} and {Chatbot} {Arena}},
  author    = {Zheng, Lianmin and Chiang, Wei-Lin and Sheng, Ying and Zhuang, Siyuan and Wu, Zhanghao and Zhuang, Yonghao and Lin, Zi and Li, Zhuohan and Li, Dacheng and Xing, Eric P. and Zhang, Hao and Gonzalez, Joseph E. and Stoica, Ion},
  booktitle = {Advances in Neural Information Processing Systems 36 (NeurIPS 2023), Datasets and Benchmarks Track},
  year      = {2023}
}

@inproceedings{deutsch2022limitations,
  title     = {On the Limitations of Reference-Free Evaluations of Generated Text},
  author    = {Deutsch, Daniel and Dror, Rotem and Roth, Dan},
  booktitle = {Proceedings of the 2022 Conference on Empirical Methods in Natural Language Processing (EMNLP)},
  year      = {2022},
  address   = {Abu Dhabi, United Arab Emirates},
  publisher = {Association for Computational Linguistics}
}

@article{reiter2018structured,
  title={A structured review of the validity of BLEU},
  author={Reiter, Ehud},
  journal={Computational Linguistics},
  volume={44},
  number={3},
  pages={393--401},
  year={2018}
}

@inproceedings{novikova2017we,
  title={Why we need new evaluation metrics for NLG},
  author={Novikova, Jekaterina and Du{\v{s}}ek, Ond{\v{r}}ej and Curry, Amanda Cercas and Rieser, Verena},
  booktitle={Proceedings of the 2017 conference on empirical methods in natural language processing},
  pages={2241--2252},
  year={2017}
}

@inproceedings{liu2016not,
  title={How not to evaluate your dialogue system: An empirical study of unsupervised evaluation metrics for dialogue response generation},
  author={Liu, Chia-Wei and Lowe, Ryan and Serban, Iulian Vlad and Noseworthy, Mike and Charlin, Laurent and Pineau, Joelle},
  booktitle={Proceedings of the 2016 conference on empirical methods in natural language processing},
  pages={2122--2132},
  year={2016}
}

@article{everingham2010pascal,
  title={The {PASCAL} {V}isual {O}bject {C}lasses ({VOC}) {C}hallenge},
  author={Everingham, Mark and Van Gool, Luc and Williams, Christopher K.~I. and Winn, John and Zisserman, Andrew},
  journal={International Journal of Computer Vision},
  volume={88},
  number={2},
  pages={303--338},
  year={2010},
  publisher={Springer}
}

@article{gardner2013classification,
  title={A classification of genre families in university student writing},
  author={Gardner, Sheena and Nesi, Hilary},
  journal={Applied linguistics},
  volume={34},
  number={1},
  pages={25--52},
  year={2013},
  publisher={Oxford University Press}
}

@inproceedings{gomez2023confederacy,
  title={A confederacy of models: a comprehensive evaluation of LLMs on creative writing},
  author={G{\'o}mez-Rodr{\'\i}guez, Carlos and Williams, Paul},
  booktitle={Findings of the Association for Computational Linguistics: EMNLP 2023},
  pages={14504--14528},
  year={2023}
}

@inproceedings{chakrabarty2024art,
  title={Art or artifice? large language models and the false promise of creativity},
  author={Chakrabarty, Tuhin and Laban, Philippe and Agarwal, Divyansh and Muresan, Smaranda and Wu, Chien-Sheng},
  booktitle={Proceedings of the 2024 CHI Conference on Human Factors in Computing Systems},
  pages={1--34},
  year={2024}
}

@article{straub1996concept,
  title={The concept of control in teacher response: Defining the varieties of “directive” and “facilitative” commentary},
  author={Straub, Richard},
  journal={College Composition \& Communication},
  volume={47},
  number={2},
  pages={223--251},
  year={1996},
  publisher={NCTE}
}

@article{hou2025improve,
  title={Improve llm-based automatic essay scoring with linguistic features},
  author={Hou, Zhaoyi Joey and Ciuba, Alejandro and Li, Xiang Lorraine},
  journal={arXiv preprint arXiv:2502.09497},
  year={2025}
}

@inproceedings{behzad2024leaf,
  title={LEAF: Language learners’ English essays and feedback corpus},
  author={Behzad, Shabnam and Kashefi, Omid and Somasundaran, Swapna},
  booktitle={Proceedings of the 2024 Conference of the North American Chapter of the Association for Computational Linguistics: Human Language Technologies (Volume 2: Short Papers)},
  pages={433--442},
  year={2024}
}

@article{zyska2026expos,
  title={Expos\'ia: Academic Writing Assessment of Expos\'es and Peer Feedback},
  author={Zyska, Dennis and Rozovskaya, Alla and Kuznetsov, Ilia and Gurevych, Iryna},
  journal={arXiv preprint arXiv:2601.06536},
  year={2026}
}

@article{kuhn1955hungarian,
  title={The Hungarian method for the assignment problem},
  author={Kuhn, Harold W},
  journal={Naval research logistics quarterly},
  volume={2},
  number={1-2},
  pages={83--97},
  year={1955},
  publisher={Wiley Online Library}
}

@inproceedings{luo2005coreference,
  title={On coreference resolution performance metrics},
  author={Luo, Xiaoqiang},
  booktitle={Proceedings of human language technology conference and conference on empirical methods in natural language processing},
  pages={25--32},
  year={2005}
}

@article{song2020mpnet,
  title={Mpnet: Masked and permuted pre-training for language understanding},
  author={Song, Kaitao and Tan, Xu and Qin, Tao and Lu, Jianfeng and Liu, Tie-Yan},
  journal={Advances in neural information processing systems},
  volume={33},
  pages={16857--16867},
  year={2020}
}

@article{wang2020minilm,
  title={Minilm: Deep self-attention distillation for task-agnostic compression of pre-trained transformers},
  author={Wang, Wenhui and Wei, Furu and Dong, Li and Bao, Hangbo and Yang, Nan and Zhou, Ming},
  journal={Advances in neural information processing systems},
  volume={33},
  pages={5776--5788},
  year={2020}
}

@inproceedings{popovic2015chrf,
  title={chrF: character n-gram F-score for automatic MT evaluation},
  author={Popovi{\'c}, Maja},
  booktitle={Proceedings of the tenth workshop on statistical machine translation},
  pages={392--395},
  year={2015}
}

@inproceedings{agirre2012semeval,
  title={Semeval-2012 task 6: A pilot on semantic textual similarity. in* sem 2012: The first joint conference on lexical and computational semantics--volume 1: Proceedings of the main conference and the shared task, and volume 2: Proceedings of the sixth international workshop on semantic evaluation (semeval 2012)},
  author={Agirre, Eneko and Cer, Daniel and Diab, Mona and Gonzalez-Agirre, Aitor},
  booktitle={Proceedings of the Sixth International Workshop on Semantic Evaluation (SemEval 2012), Montr{\'e}al, QC, Canada},
  pages={7--8},
  year={2012}
}

@article{majumder2016semantic,
  title={Semantic textual similarity methods, tools, and applications: A survey},
  author={Majumder, Goutam and Pakray, Partha and Gelbukh, Alexander and Pinto, David},
  journal={Computaci{\'o}n y Sistemas},
  volume={20},
  number={4},
  pages={647--665},
  year={2016},
  publisher={Instituto Polit{\'e}cnico Nacional, Centro de Investigaci{\'o}n en Computaci{\'o}n}
}

@inproceedings{bar2012ukp,
  title={Ukp: Computing semantic textual similarity by combining multiple content similarity measures},
  author={B{\"a}r, Daniel and Biemann, Chris and Gurevych, Iryna and Zesch, Torsten},
  booktitle={* SEM 2012: The First Joint Conference on Lexical and Computational Semantics--Volume 1: Proceedings of the main conference and the shared task, and Volume 2: Proceedings of the Sixth International Workshop on Semantic Evaluation (SemEval 2012)},
  pages={435--440},
  year={2012}
}

@article{pevzner2002critique,
  title={A critique and improvement of an evaluation metric for text segmentation},
  author={Pevzner, Lev and Hearst, Marti A},
  journal={Computational Linguistics},
  volume={28},
  number={1},
  pages={19--36},
  year={2002},
  publisher={MIT Press One Rogers Street, Cambridge, MA 02142-1209, USA journals-info~…}
}

@article{stab2017parsing,
  title={Parsing argumentation structures in persuasive essays},
  author={Stab, Christian and Gurevych, Iryna},
  journal={Computational Linguistics},
  volume={43},
  number={3},
  pages={619--659},
  year={2017},
  publisher={MIT Press One Rogers Street, Cambridge, MA 02142-1209, USA journals-info~…}
}

@article{hua2019argument,
  title={Argument mining for understanding peer reviews},
  author={Hua, Xinyu and Nikolov, Mitko and Badugu, Nikhil and Wang, Lu},
  journal={arXiv preprint arXiv:1903.10104},
  year={2019}
}

@article{wang2018toward,
  title={Toward fast and accurate neural discourse segmentation},
  author={Wang, Yizhong and Li, Sujian and Yang, Jingfeng},
  journal={arXiv preprint arXiv:1808.09147},
  year={2018}
}

@article{li2022survey,
  title={A survey of discourse parsing},
  author={Li, Jiaqi and Liu, Ming and Qin, Bing and Liu, Ting},
  journal={Frontiers of Computer Science},
  volume={16},
  number={5},
  pages={165329},
  year={2022},
  publisher={Springer}
}

@inproceedings{soricut2003sentence,
  title={Sentence level discourse parsing using syntactic and lexical information},
  author={Soricut, Radu and Marcu, Daniel},
  booktitle={Proceedings of the 2003 Human Language Technology Conference of the North American Chapter of the Association for Computational Linguistics},
  pages={228--235},
  year={2003}
}

@book{marcu2000theory,
  title={The theory and practice of discourse parsing and summarization},
  author={Marcu, Daniel},
  year={2000},
  publisher={MIT press}
}

@inproceedings{li2014recursive,
  title={Recursive deep models for discourse parsing},
  author={Li, Jiwei and Li, Rumeng and Hovy, Eduard},
  booktitle={Proceedings of the 2014 conference on empirical methods in natural language processing (EMNLP)},
  pages={2061--2069},
  year={2014}
}

@article{zou2024investigating,
  title={Investigating students’ uptake of teacher-and ChatGPT-generated feedback in EFL writing: A comparison study},
  author={Zou, Shaoyan and Guo, Kai and Wang, Jun and Liu, Yu},
  journal={Computer Assisted Language Learning},
  pages={1--30},
  year={2024},
  publisher={Taylor \& Francis}
}

@inproceedings{yen2020decipher,
  title={Decipher: an interactive visualization tool for interpreting unstructured design feedback from multiple providers},
  author={Yen, Yu-Chun Grace and Kim, Joy O and Bailey, Brian P},
  booktitle={Proceedings of the 2020 CHI Conference on Human Factors in Computing Systems},
  pages={1--13},
  year={2020}
}

@article{lyu2024steps,
  title={Steps to implementation: the role of peer feedback inner structure on feedback implementation},
  author={Lyu, Qianru and Chen, Wenli and Su, Junzhu and Heng, Kok Hui John Gerard},
  journal={Assessment \& Evaluation in Higher Education},
  volume={49},
  number={4},
  pages={572--585},
  year={2024},
  publisher={Taylor \& Francis}
}

@article{wu2020feedback,
  title={From feedback to revisions: Effects of feedback features and perceptions},
  author={Wu, Yong and Schunn, Christian D},
  journal={Contemporary Educational Psychology},
  volume={60},
  pages={101826},
  year={2020},
  publisher={Elsevier}
}

@inproceedings{lyu2023faithful,
  title={Faithful chain-of-thought reasoning},
  author={Lyu, Qing and Havaldar, Shreya and Stein, Adam and Zhang, Li and Rao, Delip and Wong, Eric and Apidianaki, Marianna and Callison-Burch, Chris},
  booktitle={The 13th International Joint Conference on Natural Language Processing and the 3rd Conference of the Asia-Pacific Chapter of the Association for Computational Linguistics (IJCNLP-AACL 2023)},
  year={2023}
}

@article{wei2022chain,
  title={Chain-of-thought prompting elicits reasoning in large language models},
  author={Wei, Jason and Wang, Xuezhi and Schuurmans, Dale and Bosma, Maarten and Xia, Fei and Chi, Ed and Le, Quoc V and Zhou, Denny and others},
  journal={Advances in neural information processing systems},
  volume={35},
  pages={24824--24837},
  year={2022}
}

@article{qwen2025qwen25,
  title  = {Qwen2.5 Technical Report},
  author = {Qwen and Yang, An and Yang, Baosong and Zhang, Beichen and Hui, Binyuan and Zheng, Bo and Yu, Bowen and Li, Chengyuan and Liu, Dayiheng and Huang, Fei and Wei, Haoran and Lin, Huan and Yang, Jian and Tu, Jianhong and Zhang, Jianwei and Yang, Jianxin and Yang, Jiaxi and Zhou, Jingren and Lin, Junyang and Dang, Kai and Lu, Keming and Bao, Keqin and Yang, Kexin and Yu, Le and Li, Mei and Xue, Mingfeng and Zhang, Pei and Zhu, Qin and Men, Rui and Lin, Runji and Li, Tianhao and Tang, Tianyi and Xia, Tingyu and Ren, Xingzhang and Ren, Xuancheng and Fan, Yang and Su, Yang and Zhang, Yichang and Wan, Yu and Liu, Yuqiong and Cui, Zeyu and Zhang, Zhenru and Qiu, Zihan},
  journal = {arXiv preprint arXiv:2412.15115},
  year   = {2025}
}

@inproceedings{dahlmeier2013building,
  title={Building a large annotated corpus of learner English: The NUS corpus of learner English},
  author={Dahlmeier, Daniel and Ng, Hwee Tou and Wu, Siew Mei},
  booktitle={Proceedings of the eighth workshop on innovative use of NLP for building educational applications},
  pages={22--31},
  year={2013}
}

@article{lipnevich2009really,
  title={“I really need feedback to learn:” students’ perspectives on the effectiveness of the differential feedback messages},
  author={Lipnevich, Anastasiya A and Smith, Jeffrey K},
  journal={Educational Assessment, Evaluation and Accountability},
  volume={21},
  number={4},
  pages={347--367},
  year={2009},
  publisher={Springer}
}

@article{lee2015cityu,
  title={CityU corpus of essay drafts of English language learners: a corpus of textual revision in second language writing},
  author={Lee, John and Yeung, Chak Yan and Zeldes, Amir and Reznicek, Marc and L{\"u}deling, Anke and Webster, Jonathan},
  journal={Language Resources and Evaluation},
  volume={49},
  number={3},
  pages={659--683},
  year={2015},
  publisher={Springer}
}

@misc{jiang2023mistral7b,
      title={Mistral 7B}, 
      author={Albert Q. Jiang and Alexandre Sablayrolles and Arthur Mensch and Chris Bamford and Devendra Singh Chaplot and Diego de las Casas and Florian Bressand and Gianna Lengyel and Guillaume Lample and Lucile Saulnier and Lélio Renard Lavaud and Marie-Anne Lachaux and Pierre Stock and Teven Le Scao and Thibaut Lavril and Thomas Wang and Timothée Lacroix and William El Sayed},
      year={2023},
      eprint={2310.06825},
      archivePrefix={arXiv},
      primaryClass={cs.CL},
      url={https://arxiv.org/abs/2310.06825}, 
}

@article{guo2025deepseek,
  title={Deepseek-r1: Incentivizing reasoning capability in llms via reinforcement learning},
  author={Guo, Daya and Yang, Dejian and Zhang, Haowei and Song, Junxiao and Zhang, Ruoyu and Xu, Runxin and Zhu, Qihao and Ma, Shirong and Wang, Peiyi and Bi, Xiao and others},
  journal={arXiv preprint arXiv:2501.12948},
  year={2025}
}

@article{zhang2019bertscore,
  title={Bertscore: Evaluating text generation with bert},
  author={Zhang, Tianyi and Kishore, Varsha and Wu, Felix and Weinberger, Kilian Q and Artzi, Yoav},
  journal={arXiv preprint arXiv:1904.09675},
  year={2019}
}

@article{yuan2021bartscore,
  title={Bartscore: Evaluating generated text as text generation},
  author={Yuan, Weizhe and Neubig, Graham and Liu, Pengfei},
  journal={Advances in neural information processing systems},
  volume={34},
  pages={27263--27277},
  year={2021}
}

@inproceedings{sellam2020bleurt,
  title={BLEURT: Learning robust metrics for text generation},
  author={Sellam, Thibault and Das, Dipanjan and Parikh, Ankur},
  booktitle={Proceedings of the 58th annual meeting of the association for computational linguistics},
  pages={7881--7892},
  year={2020}
}

@inproceedings{lin2004rouge,
  title={Rouge: A package for automatic evaluation of summaries},
  author={Lin, Chin-Yew},
  booktitle={Text summarization branches out},
  pages={74--81},
  year={2004}
}

@inproceedings{papineni2002bleu,
  title={Bleu: a method for automatic evaluation of machine translation},
  author={Papineni, Kishore and Roukos, Salim and Ward, Todd and Zhu, Wei-Jing},
  booktitle={Proceedings of the 40th annual meeting of the Association for Computational Linguistics},
  pages={311--318},
  year={2002}
}

@inproceedings{mathias2018asap++,
  title={ASAP++: Enriching the ASAP automated essay grading dataset with essay attribute scores},
  author={Mathias, Sandeep and Bhattacharyya, Pushpak},
  booktitle={Proceedings of the eleventh international conference on language resources and evaluation (LREC 2018)},
  year={2018}
}

@article{celikyilmaz2020evaluation,
  title={Evaluation of text generation: A survey},
  author={Celikyilmaz, Asli and Clark, Elizabeth and Gao, Jianfeng},
  journal={arXiv preprint arXiv:2006.14799},
  year={2020}
}

@article{crossley2024large,
  title={A large-scale corpus for assessing written argumentation: PERSUADE 2.0},
  author={Crossley, Scott A and Tian, Y and Baffour, P and Franklin, Abigail and Benner, Margaret and Boser, Ulrich},
  journal={Assessing Writing},
  volume={61},
  pages={100865},
  year={2024},
  publisher={Elsevier}
}

@article{kashefi2022argrewrite,
  title={Argrewrite v. 2: an annotated argumentative revisions corpus},
  author={Kashefi, Omid and Afrin, Tazin and Dale, Meghan and Olshefski, Christopher and Godley, Amanda and Litman, Diane and Hwa, Rebecca},
  journal={Language Resources and Evaluation},
  volume={56},
  number={3},
  pages={881--915},
  year={2022},
  publisher={Springer}
}

@inproceedings{samsi2023words,
  title={From words to watts: Benchmarking the energy costs of large language model inference},
  author={Samsi, Siddharth and Zhao, Dan and McDonald, Joseph and Li, Baolin and Michaleas, Adam and Jones, Michael and Bergeron, William and Kepner, Jeremy and Tiwari, Devesh and Gadepally, Vijay},
  booktitle={2023 IEEE High Performance Extreme Computing Conference (HPEC)},
  pages={1--9},
  year={2023},
  organization={IEEE}
}

@article{fernandez2025energy,
  title={Energy considerations of large language model inference and efficiency optimizations},
  author={Fernandez, Jared and Na, Clara and Tiwari, Vashisth and Bisk, Yonatan and Luccioni, Sasha and Strubell, Emma},
  journal={arXiv preprint arXiv:2504.17674},
  year={2025}
}

@article{huang2025survey,
  title={A survey on hallucination in large language models: Principles, taxonomy, challenges, and open questions},
  author={Huang, Lei and Yu, Weijiang and Ma, Weitao and Zhong, Weihong and Feng, Zhangyin and Wang, Haotian and Chen, Qianglong and Peng, Weihua and Feng, Xiaocheng and Qin, Bing and others},
  journal={ACM Transactions on Information Systems},
  volume={43},
  number={2},
  pages={1--55},
  year={2025},
  publisher={ACM New York, NY}
}

@inproceedings{endignoux2016caradoc,
  title={Caradoc: A pragmatic approach to pdf parsing and validation},
  author={Endignoux, Guillaume and Levillain, Olivier and Migeon, Jean-Yves},
  booktitle={2016 IEEE Security and Privacy Workshops (SPW)},
  pages={126--139},
  year={2016},
  organization={Ieee}
}

@inproceedings{ke2019give,
  title={Give me more feedback II: Annotating thesis strength and related attributes in student essays},
  author={Ke, Zixuan and Inamdar, Hrishikesh and Lin, Hui and Ng, Vincent},
  booktitle={Proceedings of the 57th annual meeting of the association for computational linguistics},
  pages={3994--4004},
  year={2019}
}

@inproceedings{ke2019automated,
  title={Automated Essay Scoring: A Survey of the State of the Art.},
  author={Ke, Zixuan and Ng, Vincent},
  booktitle={IJCAI},
  volume={19},
  pages={6300--6308},
  year={2019}
}

@article{ramesh2022automated,
  title={An automated essay scoring systems: a systematic literature review},
  author={Ramesh, Dadi and Sanampudi, Suresh Kumar},
  journal={Artificial Intelligence Review},
  volume={55},
  number={3},
  pages={2495--2527},
  year={2022},
  publisher={Springer}
}

@article{graham2019changing,
  title={Changing how writing is taught},
  author={Graham, Steve},
  journal={Review of Research in Education},
  volume={43},
  number={1},
  pages={277--303},
  year={2019},
  publisher={SAGE Publications Sage CA: Los Angeles, CA}
}

@article{applebee2011ej,
  title={EJ Extra: A snapshot of writing instruction in middle schools and high schools [free access]},
  author={Applebee, Arthur N and Langer, Judith A},
  journal={English journal},
  volume={100},
  number={6},
  pages={14--27},
  year={2011},
  publisher={ncte. org}
}

@article{kluger1996effects,
  title={The effects of feedback interventions on performance: a historical review, a meta-analysis, and a preliminary feedback intervention theory.},
  author={Kluger, Avraham N and DeNisi, Angelo},
  journal={Psychological bulletin},
  volume={119},
  number={2},
  pages={254},
  year={1996},
  publisher={American Psychological Association}
}

@article{carless2018development,
  title={The development of student feedback literacy: enabling uptake of feedback},
  author={Carless, David and Boud, David},
  journal={Assessment \& Evaluation in Higher Education},
  volume={43},
  number={8},
  pages={1315--1325},
  year={2018},
  publisher={Taylor \& Francis}
}

@article{jia2022insta,
  title={Insta-Reviewer: A Data-Driven Approach for Generating Instant Feedback on Students' Project Reports.},
  author={Jia, Qinjin and Young, Mitchell and Xiao, Yunkai and Cui, Jialin and Liu, Chengyuan and Rashid, Parvez and Gehringer, Edward},
  journal={International Educational Data Mining Society},
  year={2022},
  publisher={ERIC}
}

@article{ahea2016value,
  title={The Value and Effectiveness of Feedback in Improving Students' Learning and Professionalizing Teaching in Higher Education.},
  author={Ahea, Md Mamoon-Al-Bashir and Ahea, Md Rezaul Kabir and Rahman, Ismat},
  journal={Journal of Education and Practice},
  volume={7},
  number={16},
  pages={38--41},
  year={2016},
  publisher={ERIC}
}

@article{banihashem2024feedback,
  title={Feedback sources in essay writing: peer-generated or AI-generated feedback?},
  author={Banihashem, Seyyed Kazem and Kerman, Nafiseh Taghizadeh and Noroozi, Omid and Moon, Jewoong and Drachsler, Hendrik},
  journal={International Journal of Educational Technology in Higher Education},
  volume={21},
  number={1},
  pages={23},
  year={2024},
  publisher={Springer}
}

@article{hattie2007power,
  title={The power of feedback},
  author={Hattie, John and Timperley, Helen},
  journal={Review of educational research},
  volume={77},
  number={1},
  pages={81--112},
  year={2007},
  publisher={Sage Publications Sage CA: Thousand Oaks, CA}
}

\appendix
\section{Dataset}
\subsection{Data Collection Materials}\label{sec:data_collection}
The study was approved by the university's IRB (\S\ref{sec:ethics}). The recruitment materials distributed to students and instructors are reproduced below.
\begin{figure*}[p]
    \centering
    \includegraphics[width=0.95\textwidth,page=1]{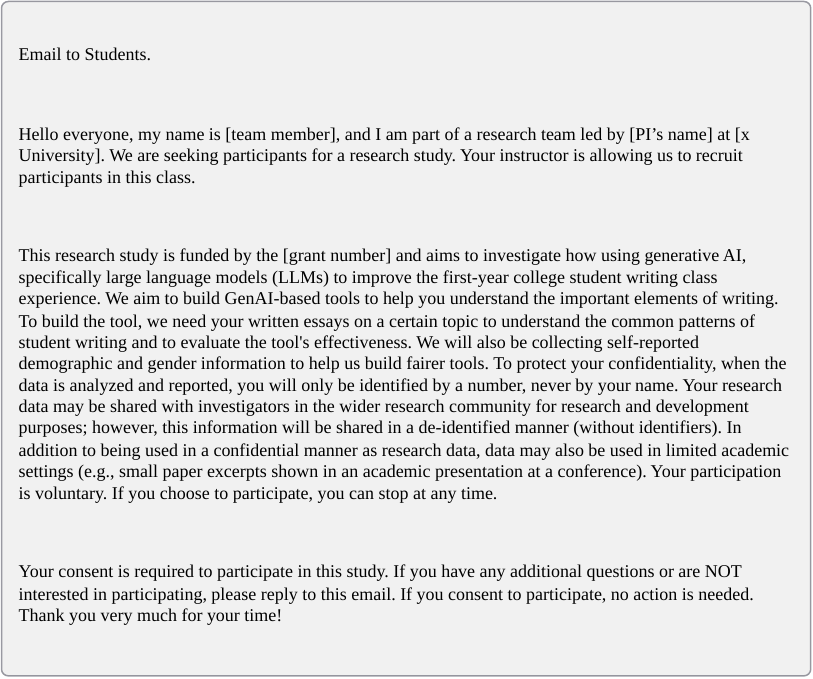}
    \caption{Recruitment email for students.}
\end{figure*}

\begin{figure*}[p]
    \centering
    \includegraphics[width=0.95\textwidth,page=1]{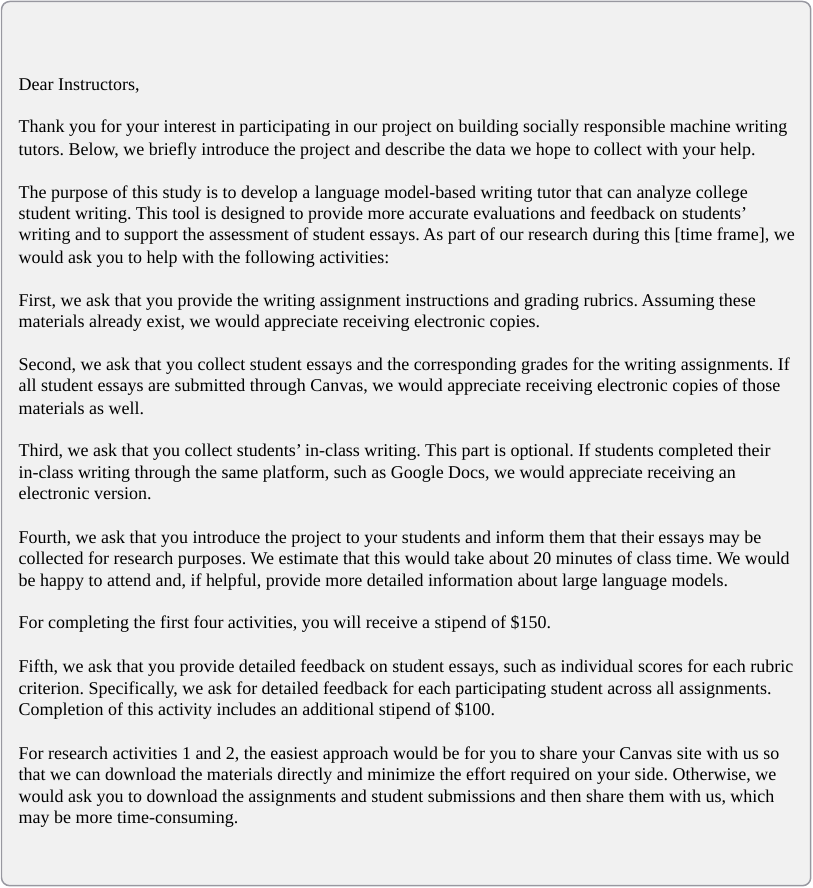}
    \caption{Recruitment email for instructor part 1.}
\end{figure*}

\begin{figure*}[p]
    \centering
    \includegraphics[width=0.95\textwidth,page=1]{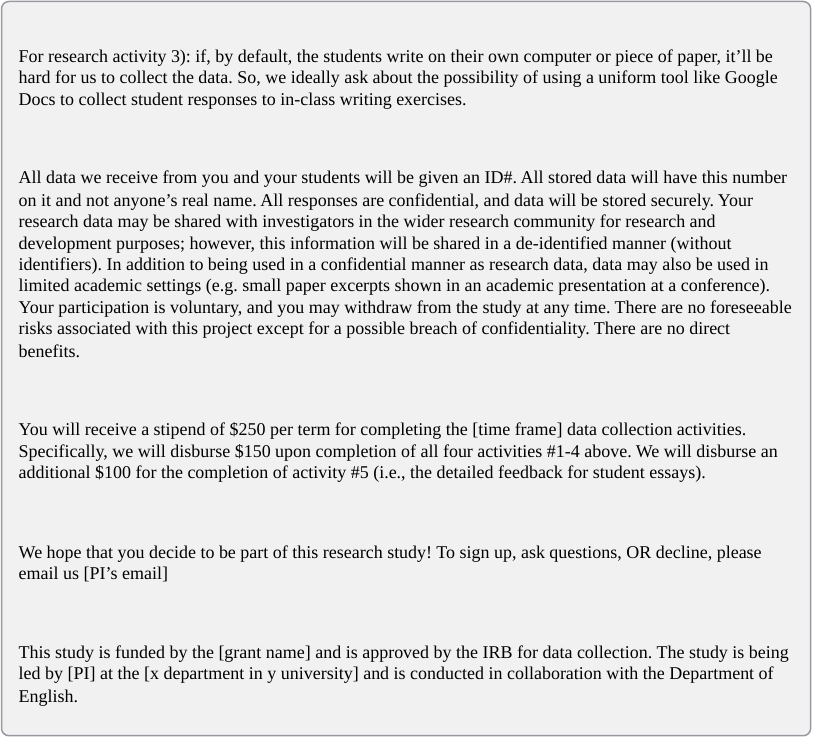}
    \caption{Recruitment email for instructor part 2.}
\end{figure*}

\subsection{Annotation Representation Details}\label{sec:annotation_details}
Paragraph-level annotations are stored within the corresponding paragraph, while document-level comments and scores are stored outside the paragraph structure.

Sticky notes are stored with their text and exact positions within the paragraph body, reflecting the original placement. Highlights are stored with their exact spans and associated comments; in our data, nearly all highlights contain instructor-written comment text. For all paragraph-level annotations we also retain the immediate left and right context of the marked span to facilitate locating the annotation within the paragraph, especially when the highlighted text is short or when a sticky note is tied to a specific local position in the essay. When instructors use non-generic highlight colors, we preserve these as structured signals. In this dataset, color usage follows a consistent instructor-defined legend:
\begin{itemize}\setlength\itemsep{0pt}\setlength\parskip{0pt}\setlength\parsep{0pt}
    \item Pink: ``great job, this is an awesome line''
    \item Green: ``there is something really interesting here \ldots expand on this point in your next draft''
    \item Yellow: ``there is some problem with this sentence''
\end{itemize}
\subsection{Parsing Details}\label{sec:parsing_details}
A central preprocessing challenge is preserving both paragraph structure and annotation anchoring across annotated PDF and \texttt{.docx} submissions. For \texttt{.docx}, we extract standard highlights and comments with their anchored spans. For PDFs, however, annotations must be reconstructed from page-level geometry, including highlight quadrilaterals and note positions, rather than from a linear text stream \cite{endignoux2016caradoc}.

\begin{figure}[t]
\centering
\small
\begin{tikzpicture}[
  node distance=5pt,
  stage/.style={
    draw, rounded corners=2pt, align=left,
    text width=0.86\columnwidth, inner sep=5pt, font=\footnotesize
  },
  arr/.style={-{Latex[length=4pt]}, draw=black!60}
]
\node[stage] (body) {\textbf{1. Body text extraction} \\ Deterministic. Extracts every character in the box.};
\node[stage, below=of body] (hl) {\textbf{2. Highlight anchoring} \\ Quads $\rightarrow$ covered text $\rightarrow$ matched span; comment stored.};
\node[stage, below=of hl] (color) {\textbf{3. Color recovery} \\ Local render $\rightarrow$ HEX $\rightarrow$ named color.};
\node[stage, below=of color] (note) {\textbf{4. Sticky-note anchoring} \\ Coordinate-radius context $\rightarrow$ located in body $\rightarrow$ injected.};
\node[stage, below=of note] (para) {\textbf{5. Paragraph structure} \\ Inferred from inter-block spacing (visual cues).};
\node[stage, below=of para] (clean) {\textbf{6. Identifier removal} \\ Positional headers/footers; regex metadata; human-checked.};
\draw[arr] (body) -- (hl);
\draw[arr] (hl) -- (color);
\draw[arr] (color) -- (note);
\draw[arr] (note) -- (para);
\draw[arr] (para) -- (clean);
\end{tikzpicture}
\caption{Stages of the deterministic PDF parsing pipeline (\S\ref{sec:parsing_details}). Each annotated submission is converted to a unified JSON representation preserving paragraph segmentation and annotation anchoring.}
\label{fig:parsing-pipeline}
\end{figure}
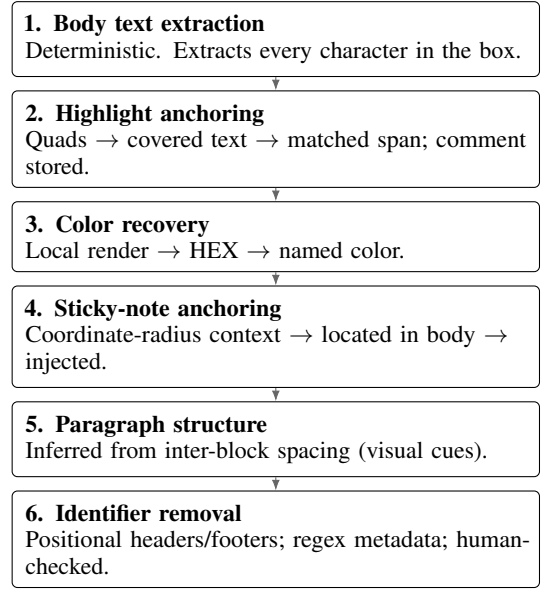

Naive PDF text extraction can interleave headers and footers and scramble paragraph boundaries, which are often conveyed by layout rather than explicit markup. We also explored LLM-based re-paragraphization of extracted text, but observed deviations from the source text, including missing or added tokens and altered boundaries, consistent with known hallucination behavior \cite{huang2025survey}; it also incurs nontrivial computational cost \cite{fernandez2025energy, samsi2023words}. We therefore implemented a deterministic parser that converts annotated PDF and \texttt{.docx} submissions into a unified JSON format while preserving paragraph segmentation and annotation anchoring. We release this parser with the dataset. The remainder of this section describes the PDF pipeline, which is the more involved of the two; the stages are summarized in Figure~\ref{fig:parsing-pipeline}.

\paragraph{Body text extraction.}
The parser first extracts the essay body from the page-level text blocks. This stage is fully deterministic: every character present in the source text blocks is retained, and no content is dropped, rewritten, or reordered by a model. The extracted body text serves as the canonical reference string to which all subsequent annotations are anchored.

\paragraph{Highlight anchoring.}
Highlights are recovered from their page-level geometry rather than from a linear text stream. For each highlight, the parser collects the highlight quadrilaterals, determines the text covered by those quads, and matches the covered text back to the corresponding span in the extracted body, anchoring the highlight to that span. The associated highlight comment, when present, is stored alongside the anchored span. Because nearly all highlights in \ourData{} carry instructor-written comment text (\S\ref{sec:annotation_details}), this step preserves both the marked span and its commentary.

\paragraph{Highlight color recovery.}
Highlight color is not reliably available as a structured attribute, so the parser recovers it visually: it performs a localized rendering of the highlighted region, reads the rendered color, and maps the resulting HEX value to the nearest entry in our set of named highlight colors. This yields the color names used as structured signals in the corpus, following the instructor-defined legend documented in \S\ref{sec:annotation_details} (e.g., pink, green, and yellow), as illustrated in Figure~\ref{fig:dataSample}.

\paragraph{Sticky-note anchoring.}
Sticky notes are positioned by page coordinates rather than attached to a text span, so the parser anchors them by location. For each note, it identifies the surrounding text within a coordinate radius of the note's placement, locates that surrounding text within the extracted body, and injects the note at the corresponding position. This reproduces the original placement of the note within the paragraph body (\S\ref{sec:annotation_details}).

\paragraph{Paragraph structure.}
Text blocks do not necessarily correspond to paragraphs, so paragraph boundaries cannot be read directly from the block structure. Instead, the parser infers paragraph structure from visual layout cues, primarily the spacing between text blocks, recovering the paragraph segmentation that layout conveys without explicit markup.

\paragraph{Header, footer, and identifier removal.}
Recurring page elements such as top-of-page names and page numbers are removed based on their position on the page. Submission metadata such as the submission date is detected using regular-expression patterns. To ensure that student and instructor names do not leak into the released corpus, positional and regex-based removal was complemented by additional scanning tools and human inspection, consistent with the privacy procedures described in \S\ref{sec:ethics}.

\section{Guidelines}
\subsection{Feedback Segmentation Annotation Guideline}\label{sec:segmentation_guidelines}
\begin{figure*}[p]
    \centering
    \includegraphics[width=0.95\textwidth,page=1]{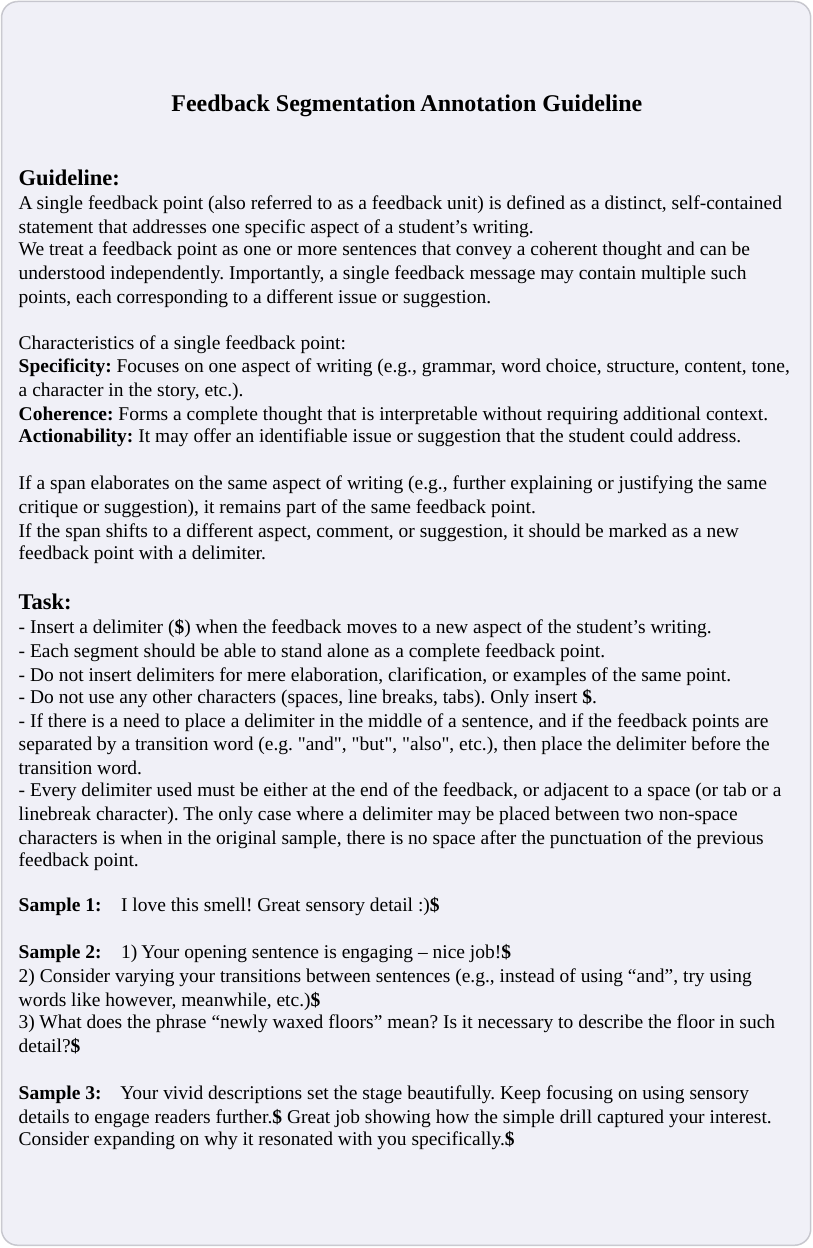}
    \caption{Feedback segmentation annotation guideline (page 1/1).}
    \label{fig:segmentation-guideline-p1}
\end{figure*}

The guideline used by annotators to segment instructor and LLM-generated feedback into feedback units (\S\ref{sec:segmentation}) is shown in Figure~\ref{fig:segmentation-guideline-p1}; segmented examples appear in Table~\ref{tab:segmentationExamples}.

\subsection{Feedback Unit Similarity Annotation Guideline}\label{sec:similarity_guidelines}
\begin{figure*}[p]
    \centering
    \includegraphics[width=0.95\textwidth,page=1]{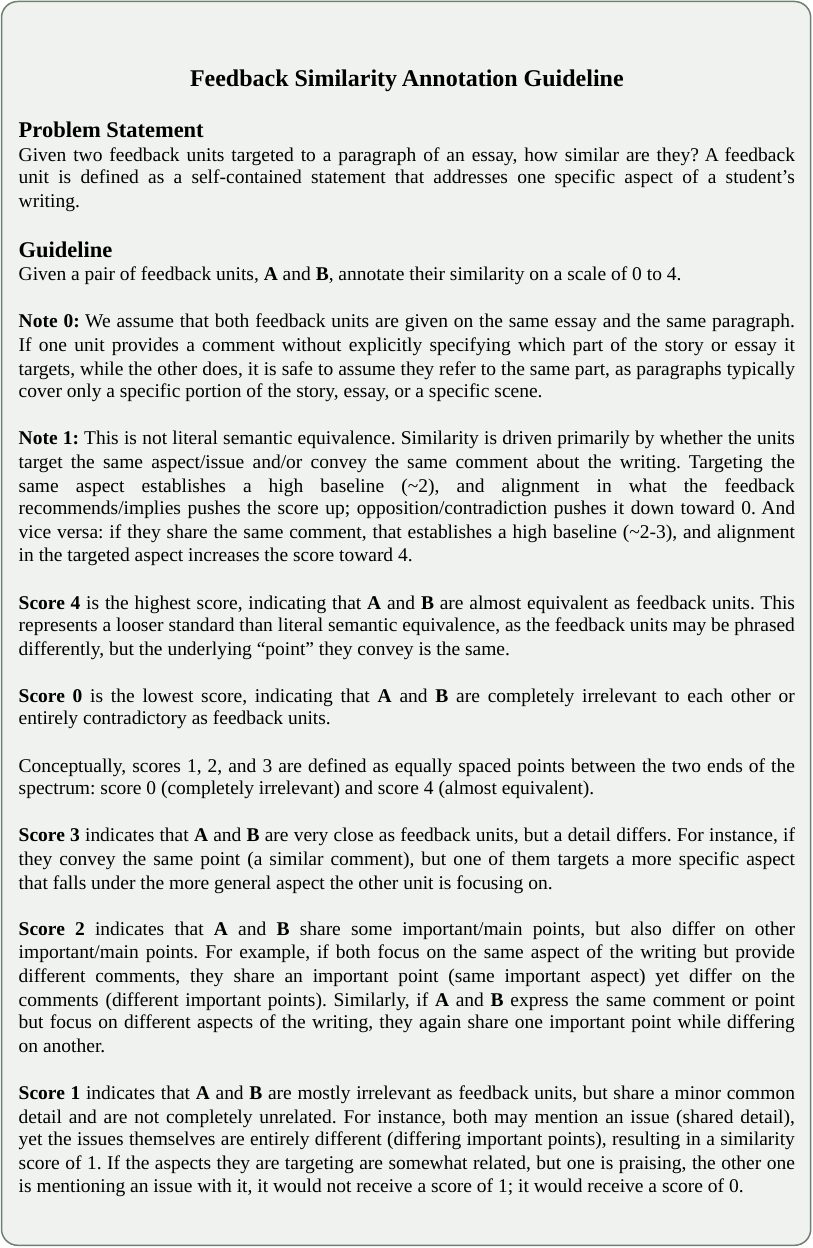}
    \caption{Feedback similarity annotation guideline (page 1/3).}
    \label{fig:similarity-guideline-p1}
\end{figure*}

\begin{figure*}[p]
    \centering
    \includegraphics[width=0.95\textwidth,page=1]{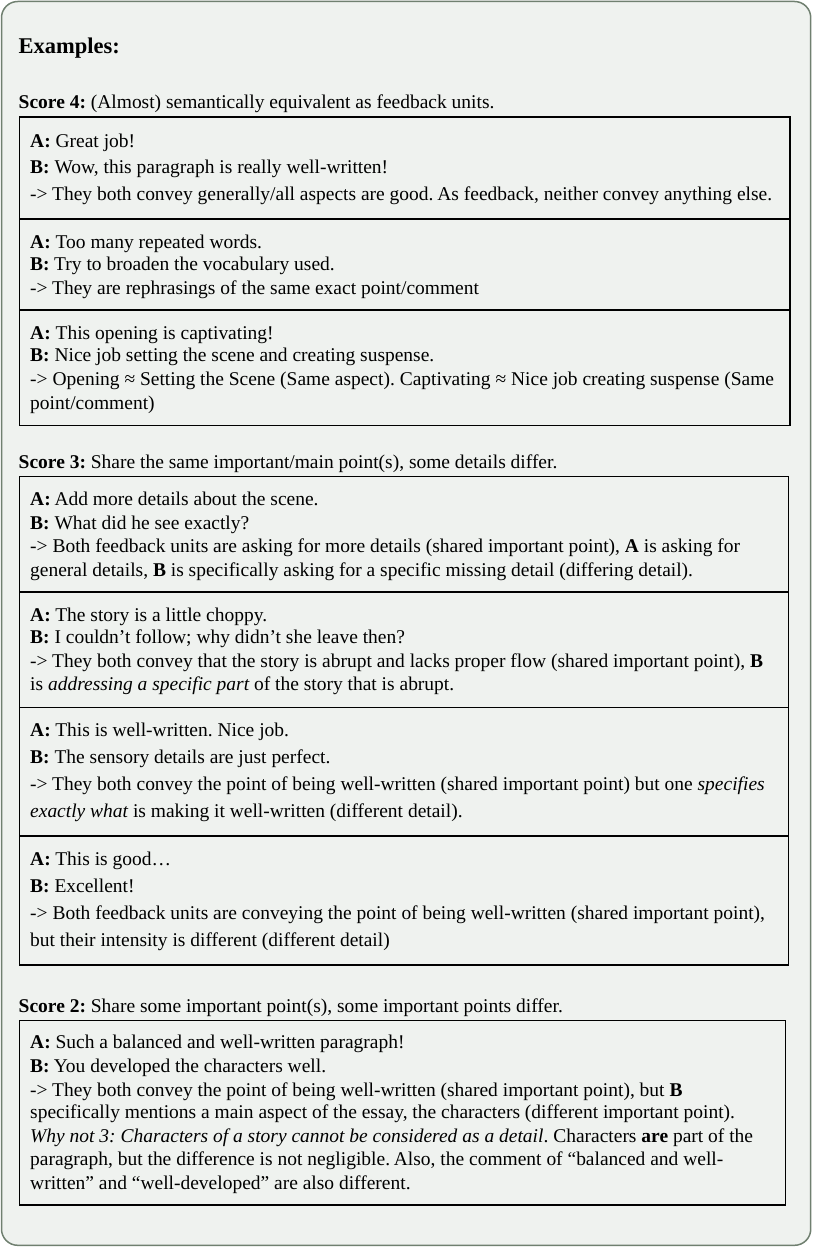}
    \caption{Feedback similarity annotation guideline (page 2/3).}
    \label{fig:similarity-guideline-p2}
\end{figure*}

\begin{figure*}[p]
    \centering
    \includegraphics[width=0.95\textwidth,page=1]{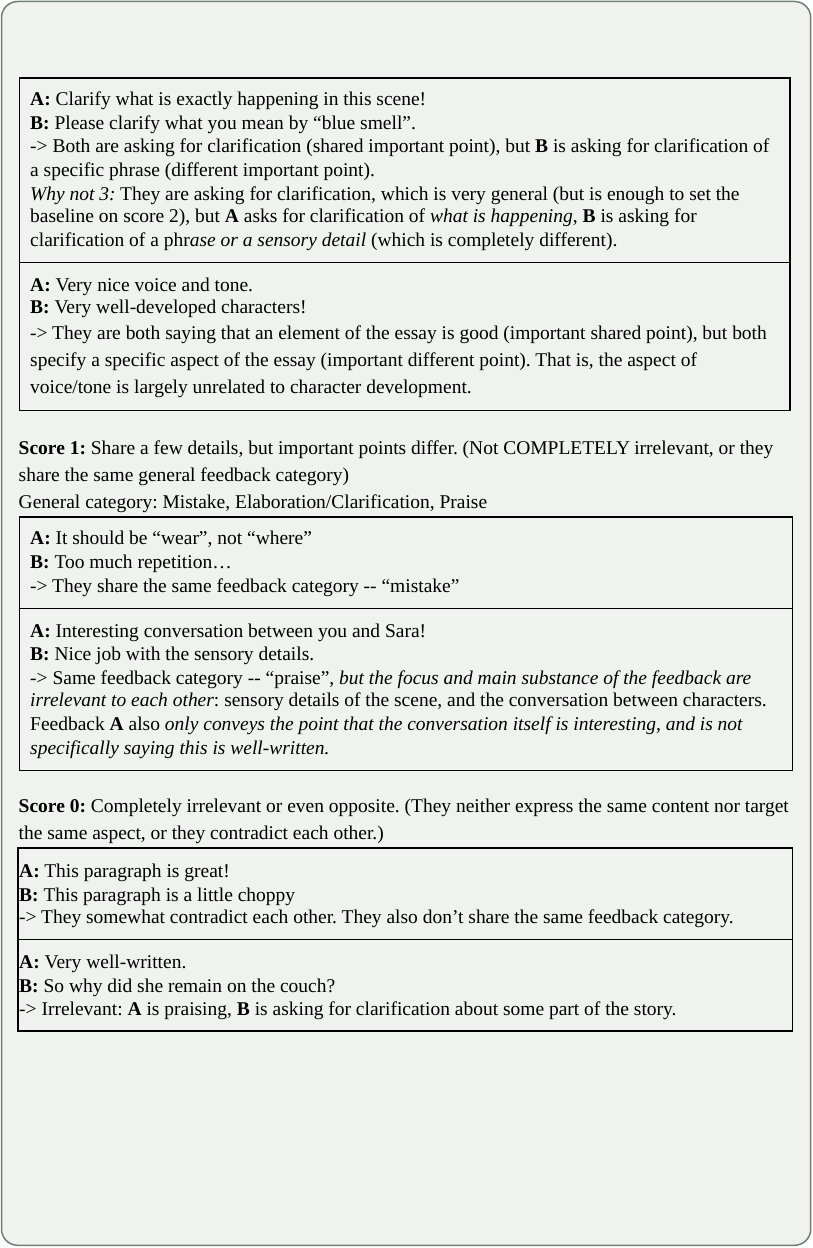}
    \caption{Feedback similarity annotation guideline (page 3/3).}
    \label{fig:similarity-guideline-p3}
\end{figure*}

The guideline used to score similarity between feedback-unit pairs on the 0--4 scale (\S\ref{sec:feedbackSimilarity}) is shown in Figures~\ref{fig:similarity-guideline-p1}--\ref{fig:similarity-guideline-p3}.

\section{Feedback Unit Similarity Pipeline} \label{sec:similarity_pipeline_appendix}
\subsection{Automatic similarity metrics.} \label{sec:automatic_similarity_metrics_appendix}
In our experiment and setting, lexical-overlap measures such as BLEU \cite{papineni2002bleu} and chrF \cite{popovic2015chrf} yield Pearson correlations below $0.1$ with annotator scores. Embedding-based approaches (MiniLM \cite{wang2020minilm}, MPNet \cite{song2020mpnet}, and BERTScore \cite{zhang2019bertscore}) perform better but still show low correlation with human scores, with Pearson correlations below $0.3$. Cosine similarity using OpenAI's embedding model \texttt{text-embedding-3-large}\footnote{\url{https://developers.openai.com/api/docs/models/text-embedding-3-large}} is the strongest automatic baseline, but its performance depends heavily on the context provided to the embedding model. Our best result ($r=0.6696$, $\rho=0.6343$) is obtained by prefixing each feedback with ``Semantic content of instructor feedback on a student-written paragraph:~'' before embedding. Even so, it remains insufficiently aligned with human judgments for reliable evaluation.

\subsection{Batch Similarity Scoring Details}\label{sec:batch_details}
\paragraph{Batched requests.} Since the similarity guideline is long, scoring every pair independently for each paragraph, draft, and experimental setting would be expensive and inefficient. We therefore score multiple pairs per request (\texttt{batch\_size}$=b$), instructing the model to evaluate each pair independently (implementation details in \S\ref{sec:batch_details}). Batching reduces the cost in inverse proportion, but since it alters the prompting setting, we reverify the agreement with the human reference. Correlation results for multiple models under the batching setup are reported in Table~\ref{tab:similarityAgreement} (\S\ref{sec:feedbackSimilarity}), where we adopt \texttt{gemini-3.1-flash-lite-preview} for the main pipeline on the strength of its agreement with human scores.

\paragraph{Batching strategy.}
The similarity guideline used for scoring feedback-unit pairs is nearly 2K tokens long (although the exact token count depends on the tokenizer), making repeated independent calls inefficient. To reduce token cost, we evaluate multiple pairs in a single request (\texttt{batch\_size}$=b$), instructing the model to assign one score to each labeled pair independently.

\paragraph{Token cost.}
Let $A$ denote the irreducible per-request token cost (system instructions, formatting, and response overhead), and $B$ the guideline length. When scoring $b$ pairs per request, the total token cost per pair can be approximated as
\[
T(b) = A + \frac{B}{b},
\]
so batching reduces the amortized guideline overhead by a factor of $O(1/b)$.

\paragraph{Effect on agreement.}
Batching slightly changes the prompting context because the model evaluates multiple pairs within the same request. We therefore verify that this does not materially affect similarity scoring. Experiments with \texttt{batch\_size}$=50$ show agreement with the human reference comparable to the single-pair setting reported in \S\ref{sec:feedbackSimilarity}, indicating that batching does not meaningfully alter model judgments.

\paragraph{Implementation notes.}
We avoided the Batch API (i.e., offline requests used to reduce API cost) in order to preserve the ability to retry malformed outputs. In auxiliary runs with \texttt{gpt-5-mini-2025-08-07}, which was not the primary model used in our main evaluation, we did not observe any syntactically invalid outputs. \texttt{gemini-3.1-flash-lite-preview}, which was used at much larger scale, produced a small number of syntactically invalid outputs, and in those cases, maximum of two retries were sufficient to obtain a syntactically valid output.

\section{Threshold-Based Cross Matching}\label{sec:appendix_threshold}
\ourMethod{}'s default cross-matching stage (\S\ref{sec:crossmatching}) aggregates similarity scores into soft precision and recall by treating each unit's full self-similarity as its maximum possible contribution. We complement this with a threshold-based variant that mirrors detection-style evaluation in computer vision \cite{everingham2010pascal}: given a threshold $\tau$ on the annotator-aligned 0--4 similarity scale, a pair counts as a match only if its similarity is at least $\tau$, and the standard hard precision, recall, and F$_1$ follow.

\paragraph{Formalization.}
Let $G=\{G_i\}_{i=1}^{|G|}$ and $P=\{P_j\}_{j=1}^{|P|}$ denote instructor and predicted feedback units for a single paragraph, with similarity scorer $\phi(G_i,P_j)\in\{0,1,2,3,4\}$. For $\tau\in\{1,2,3,4\}$, define the binarized adjacency $a_{ij}(\tau)=\mathbf{1}[\phi(G_i,P_j)\ge\tau]$ and compute the maximum bipartite matching on $a$:
\[
M^{*}_{\tau}=\arg\max_{M\in\mathcal{M}} \sum_{(G_i,P_j)\in M} a_{ij}(\tau),
\]
where $\mathcal{M}$ is the set of 1-to-1 matchings between $G$ and $P$. The match count and complementary error counts are
\[
\begin{aligned}
TP_{\tau} &= |M^{*}_{\tau}|, \\
FP_{\tau} &= |P| - TP_{\tau}, \\
FN_{\tau} &= |G| - TP_{\tau},
\end{aligned}
\]
yielding the hard metrics
\[
p_{\tau}=\frac{TP_{\tau}}{|P|},\quad r_{\tau}=\frac{TP_{\tau}}{|G|},\quad F_{1,\tau}=\frac{2\,p_{\tau}\,r_{\tau}}{p_{\tau}+r_{\tau}}.
\]

\paragraph{Interpretation of $\tau$.}
Because the similarity scale is grounded in the annotation guideline (\S\ref{sec:similarity_guidelines}), $\tau$ has a direct semantic reading: $\tau{=}1$ requires the pair to share at least a general feedback category, $\tau{=}2$ requires sharing an important point (same targeted aspect or same comment), $\tau{=}3$ requires that the units differ only in a minor detail, and $\tau{=}4$ requires near-equivalence as feedback. The threshold thus exposes how strict an ``alignment'' must be to count as feedback recovery.

\paragraph{Sensitivity to over-generation.}
In the soft variant, each additional predicted unit contributes its full self-similarity $\phi_{\max}$ to the precision denominator $\sum_j\phi(P_j,P_j)$, so a weak-but-nonzero match still inflates the denominator without comparably increasing the numerator. The threshold-based variant attenuates this: predictions affect precision only through their count, and any pair above $\tau$ receives full credit. Weak-but-above-threshold matches are therefore not partially discounted, and the per-extra-unit cost is smaller. The variant offers a less stringent view when over-generation is unavoidable; the soft variant remains preferred when fine-grained similarity differences carry signal.

\paragraph{Ranking robustness across the 74 configurations.}
Despite the qualitative difference in metric form, the relative ordering of the 74 configurations from \S\ref{sec:experiments} is preserved under thresholding: for every $\tau\in\{1,2,3,4\}$, the Spearman rank correlation between configuration F$_1$ under the soft variant and $F_{1,\tau}$ is $\rho=1.0$. The dominant findings reported in \S\ref{sec:results} -- the negative effect of verbosity, the gains from feedback-category guidance, the limited contribution of rubric inclusion, and the secondary role of few-shot prompting -- therefore do not depend on the choice of matching metric. We omit per-configuration $F_{1,\tau}$ tables, as the rank-identity result subsumes them; the implementation is released alongside the soft variant for reproduction.

\paragraph{Implementation.}
Threshold-based matching reuses the same similarity matrices produced for the soft variant; no additional model calls are required. We binarize the matrix at $\tau$ and run a bipartite matching solver on the resulting graph. Note that the Hungarian algorithm handles the general case of weighted edges (similarity scores), whereas in this setting edges are unweighted.

\section{Full Evaluation Results}\label{app:full_results}

This appendix presents per-configuration \ourMethod{} scores for all 74 settings described in \S\ref{sec:experiments}. Tables~\ref{tab:fullResultsZero},~\ref{tab:fullResultsFew}, and~\ref{tab:fullResultsCoT} report precision, recall, F$_1$, and number of generated feedback units under zero-shot, few-shot, and chain-of-thought (CoT) prompting, respectively. Each table reports both the \emph{single reference} evaluation (paragraphs with exactly one instructor feedback unit) and the \emph{full corpus} evaluation (every paragraph with at least one instructor annotation); the two regimes track each other closely on every comparison reported in \S\ref{sec:results}, and we therefore discuss only the full-corpus numbers throughout the main text. Figure~\ref{fig:effect_of_prompt} summarizes the F$_1$ effect of the prompt template across representative configurations.

\paragraph{Per-condition ablation visualizations.}
Figures~\ref{fig:effect_of_guidance}, \ref{fig:effect_of_rubric}, and \ref{fig:effect_of_fewshot} provide per-model, per-condition visualizations of the guidance, rubric, and few-shot effects summarized in Table~\ref{tab:mainResults} (\S\ref{sec:results}).

\paragraph{Prompt template.}
The default constrained prompt ($V_2$, Figure~\ref{fig:prompt-2-template}) substantially outperforms the unconstrained alternative ($V_1$, Figure~\ref{fig:prompt-1-template}) across models, conditions, and shot settings (Figure~\ref{fig:effect_of_prompt}). Across 28 matched pairs (same model, guidance, rubric, and shot), $V_2$ yields a mean F$_1$ gain of $+0.090$ over $V_1$, and wins in 27 of 28 pairs (96\%). The mechanism is verbosity: $V_1$ averages roughly 12K generated feedback units per configuration vs.\ 4.6K for $V_2$, and the resulting precision deficit overwhelms any recall benefit. This is consistent with the broader verbosity finding in \S\ref{sec:results}: the constrained prompt enforces a per-paragraph output budget that the unconstrained prompt does not.

\paragraph{Few-shot prompting under $V_1$.}
Few-shot exemplars contribute much more under the unconstrained prompt than under the constrained one. In matched comparisons, few-shot improves F$_1$ by $+0.028$ under $V_1$ but only $+0.005$ under $V_2$ (Table~\ref{tab:fullResultsFew}). This supports the interpretation in \S\ref{sec:results} that exemplars are most useful when task framing is weakest: $V_2$ already specifies the target output structure, leaving little additional ground for exemplars to cover. The few-shot gain is precision-driven across all 30 matched (zero, few) pairs: precision rises by $+0.028$ on average while recall falls slightly ($-0.004$).

\paragraph{Chain-of-thought.}
CoT prompting (Table~\ref{tab:fullResultsCoT}) was evaluated on the three small models in the zero-shot setting, using the unconstrained $V_1$ prompt. Under matched conditions, CoT improves F$_1$ by $+0.041$ over $V_1$ but underperforms $V_2$ by $-0.056$. CoT settings produce 6.6K--8.4K generated feedback units per configuration, between $V_1$ and $V_2$, and the qualitative patterns from the main results hold: Mistral is the strongest small model, feedback-category guidance contributes a small positive effect, and rubric inclusion is essentially neutral. CoT does not alter the overall story; like other settings, its position in the F$_1$ ordering is largely determined by how many feedback units it generates.

\begin{figure}[t]
\centering
\begin{tikzpicture}[scale=0.8,x=1.49cm,y=0.85cm]
\pgfplotstableread[col sep=space]{figures/effect_of_guidance.dat}\datatable
\pgfmathsetmacro{\minval}{0.216223662}
\pgfmathsetmacro{\maxval}{0.3480784335}
\pgfmathsetmacro{\midval}{0.22}   
\pgfmathsetmacro{\whiteTextAt}{65}   
\foreach \r in {0,1}{%
  \foreach \c in {0,1,2,3,4,5}{%
    \pgfplotstablegetelem{\r}{[index]\c}\of\datatable
    \pgfmathsetmacro{\val}{\pgfplotsretval}
    \pgfmathsetmacro{\isHigh}{\val>\midval ? 1 : 0}
    \ifdim\isHigh pt > 0pt
      \pgfmathsetmacro{\t}{100*(\val-\midval)/(\maxval-\midval)}
      \pgfmathsetmacro{\t}{min(max(\t,0),100)}
      \fill[cmapHigh!\t!white] (\c, 1-\r) rectangle ++(1,1);
    \else
      \pgfmathsetmacro{\t}{100*(\midval-\val)/(\midval-\minval)}
      \pgfmathsetmacro{\t}{min(max(\t,0),100)}
      \fill[cmapLow!\t!white] (\c, 1-\r) rectangle ++(1,1);
    \fi
    \pgfmathsetmacro{\useWhite}{(\isHigh>0 && \t>\whiteTextAt) ? 1 : 0}
    \ifdim\useWhite pt > 0pt
      \node[white] at (\c+0.5, 1-\r+0.5)
        {\small\pgfmathprintnumber[fixed,fixed zerofill,precision=2]{\val}};
    \else
      \node[black] at (\c+0.5, 1-\r+0.5)
        {\small\pgfmathprintnumber[fixed,fixed zerofill,precision=2]{\val}};
    \fi
  }%
}%
\node[anchor=east] at (-0, 1.5) {\small \textbf{G}};
\node[anchor=base] at (1.5, 2.25) {\small \textbf{\textit{Base}}};
\node[anchor=base] at (4.5, 2.25) {\small \textbf{R}};
\draw[thin] (0.1, 2.15) -- (2.9, 2.15);
\draw[thin] (3.1, 2.15) -- (5.9, 2.15);
\node[anchor=base] at (3, 2.45) {\small \textbf{Few}};
\draw[thin] (2.1, 2.35) -- (3.9, 2.35);
\foreach \c/\m in {0/DeepSeek, 1/Mistral, 2/Llama, 3/Qwen, 4/Llama, 5/GPT}{
  \node[anchor=base] at (\c+0.5, -0.45) {\footnotesize \m};
}
\draw[thick] (0, 2) -- (6, 2);
\draw[thick] (0, 0) -- (6, 0);
\end{tikzpicture}
\caption{Effect of guidance on F$_1$. Each column is one model under a given condition (\textit{Base}: zero-shot, no rubric; \textit{Few}: few-shot; \textit{R}: rubric); the top row adds feedback-category guidance (\textbf{G}) and the bottom row omits it. Guidance improves F$_1$ in every column. Darker cells indicate higher F$_1$.}\label{fig:effect_of_guidance}
\end{figure}

\begin{figure}[t]
\centering
\begin{subfigure}[t]{0.46\columnwidth}
\centering
\begin{tikzpicture}[scale=0.72,x=1.4cm,y=0.85cm]
\pgfplotstableread[col sep=space]{figures/effect_of_rubric.dat}\datatable
\pgfmathsetmacro{\minval}{0.203981413}
\pgfmathsetmacro{\maxval}{0.2784898858}
\pgfmathsetmacro{\midval}{0.2035}
\pgfmathsetmacro{\whiteTextAt}{65}
\foreach \r in {0,1}{%
  \foreach \c in {0,1,2}{%
    \pgfplotstablegetelem{\r}{[index]\c}\of\datatable
    \pgfmathsetmacro{\val}{\pgfplotsretval}
    \pgfmathsetmacro{\isHigh}{\val>\midval ? 1 : 0}
    \ifdim\isHigh pt > 0pt
      \pgfmathsetmacro{\t}{100*(\val-\midval)/(\maxval-\midval)}
      \pgfmathsetmacro{\t}{min(max(\t,0),100)}
      \fill[cmapHigh!\t!white] (\c, 1-\r) rectangle ++(1,1);
    \else
      \pgfmathsetmacro{\t}{100*(\midval-\val)/(\midval-\minval)}
      \pgfmathsetmacro{\t}{min(max(\t,0),100)}
      \fill[cmapLow!\t!white] (\c, 1-\r) rectangle ++(1,1);
    \fi
    \pgfmathsetmacro{\useWhite}{(\isHigh>0 && \t>\whiteTextAt) ? 1 : 0}
    \ifdim\useWhite pt > 0pt
      \node[white] at (\c+0.5, 1-\r+0.5)
        {\small\pgfmathprintnumber[fixed,fixed zerofill,precision=2]{\val}};
    \else
      \node[black] at (\c+0.5, 1-\r+0.5)
        {\small\pgfmathprintnumber[fixed,fixed zerofill,precision=2]{\val}};
    \fi
  }%
}%
\node[anchor=east] at (-0.15, 1.5) {\small \textbf{R}};
\node[anchor=east] at (-0.15, 0.5) {\small \textbf{\textit{B}}};
\node[anchor=base] at (1.0, 2.5) {\small \textbf{Few}};
\draw[thin] (0.1, 2.4) -- (2.0, 2.4);
\node[anchor=base] at (2.5, 2.3) {\small \textbf{G}};
\draw[thin] (1.0, 2.2) -- (2.9, 2.2);
\foreach \c/\m in {0/Llama, 1/Mistral, 2/Mistral}{
  \node[anchor=base] at (\c+0.5, -0.45) {\scriptsize \m};
}
\draw[thick] (0, 2) -- (3, 2);
\draw[thick] (0, 0) -- (3, 0);
\end{tikzpicture}
\caption{Rubric}\label{fig:effect_of_rubric}
\end{subfigure}
\hfill
\begin{subfigure}[t]{0.46\columnwidth}
\centering
\begin{tikzpicture}[scale=0.72,x=1.4cm,y=0.85cm]
\pgfplotstableread[col sep=space]{figures/effect_of_shot.dat}\datatable
\pgfmathsetmacro{\minval}{0.2268116649}
\pgfmathsetmacro{\maxval}{0.3685837141}
\pgfmathsetmacro{\midval}{0.2263}
\pgfmathsetmacro{\whiteTextAt}{65}
\foreach \r in {0,1}{%
  \foreach \c in {0,1,2}{%
    \pgfplotstablegetelem{\r}{[index]\c}\of\datatable
    \pgfmathsetmacro{\val}{\pgfplotsretval}
    \pgfmathsetmacro{\isHigh}{\val>\midval ? 1 : 0}
    \ifdim\isHigh pt > 0pt
      \pgfmathsetmacro{\t}{100*(\val-\midval)/(\maxval-\midval)}
      \pgfmathsetmacro{\t}{min(max(\t,0),100)}
      \fill[cmapHigh!\t!white] (\c, 1-\r) rectangle ++(1,1);
    \else
      \pgfmathsetmacro{\t}{100*(\midval-\val)/(\midval-\minval)}
      \pgfmathsetmacro{\t}{min(max(\t,0),100)}
      \fill[cmapLow!\t!white] (\c, 1-\r) rectangle ++(1,1);
    \fi
    \pgfmathsetmacro{\useWhite}{(\isHigh>0 && \t>\whiteTextAt) ? 1 : 0}
    \ifdim\useWhite pt > 0pt
      \node[white] at (\c+0.5, 1-\r+0.5)
        {\small\pgfmathprintnumber[fixed,fixed zerofill,precision=2]{\val}};
    \else
      \node[black] at (\c+0.5, 1-\r+0.5)
        {\small\pgfmathprintnumber[fixed,fixed zerofill,precision=2]{\val}};
    \fi
  }%
}%
\node[anchor=east] at (-0.15, 1.5) {\small \textbf{F}};
\node[anchor=east] at (-0.15, 0.5) {\small \textbf{Z}};
\node[anchor=base] at (1.0, 2.45) {\small \textbf{G}};
\draw[thin] (0.1, 2.35) -- (2.0, 2.35);
\node[anchor=base] at (2.5, 2.3) {\small \textbf{R}};
\draw[thin] (1.0, 2.2) -- (2.9, 2.2);
\foreach \c/\m in {0/Llama, 1/Qwen, 2/GPT}{
  \node[anchor=base] at (\c+0.5, -0.45) {\scriptsize \m};
}
\draw[thick] (0, 2) -- (3, 2);
\draw[thick] (0, 0) -- (3, 0);
\end{tikzpicture}
\caption{Few-shot}\label{fig:effect_of_fewshot}
\end{subfigure}
\caption{Effect of rubric and few-shot prompting on F$_1$, each panel on its own color scale (darker = higher F$_1$). Columns indicate a model and a generation setting; rows toggle the studied factor. \textbf{(a)} Rubric present (\textbf{R}, the assignment rubric) vs.\ absent (\textit{B}, base), across few-shot (\emph{Few}), few-shot-with-guidance, and guidance (\emph{G}, feedback categories in the prompt). \textbf{(b)} Few-shot (\textbf{F}) vs.\ zero-shot (\textbf{Z}), across guidance, guidance-with-rubric, and rubric; overlapping header lines mark the column with both settings. Both factors shift F$_1$ only slightly and inconsistently, unlike the uniform gain from guidance (Figure~\ref{fig:effect_of_guidance}).}\end{figure}

\begin{figure}[t]
\centering
\begin{tikzpicture}[scale=0.85,x=1.65cm,y=0.7cm]
\pgfplotstableread[col sep=space]{figures/effect_of_prompt.dat}\datatable
\pgfmathsetmacro{\minval}{0.126223662}
\pgfmathsetmacro{\maxval}{0.3580784335}
\pgfmathsetmacro{\midval}{0.16}
\pgfmathsetmacro{\whiteTextAt}{65}   
\foreach \r in {0,1}{%
  \foreach \c in {0,1,2,3,4}{%
    \pgfplotstablegetelem{\r}{[index]\c}\of\datatable
    \pgfmathsetmacro{\val}{\pgfplotsretval}
    \pgfmathsetmacro{\isHigh}{\val>\midval ? 1 : 0}
    \ifdim\isHigh pt > 0pt
      \pgfmathsetmacro{\t}{100*(\val-\midval)/(\maxval-\midval)}
      \pgfmathsetmacro{\t}{min(max(\t,0),100)}
      \fill[cmapHigh!\t!white] (\c, 1-\r) rectangle ++(1,1);
    \else
      \pgfmathsetmacro{\t}{100*(\midval-\val)/(\midval-\minval)}
      \pgfmathsetmacro{\t}{min(max(\t,0),100)}
      \fill[cmapLow!\t!white] (\c, 1-\r) rectangle ++(1,1);
    \fi
    \pgfmathsetmacro{\useWhite}{(\isHigh>0 && \t>\whiteTextAt) ? 1 : 0}
    \ifdim\useWhite pt > 0pt
      \node[white] at (\c+0.5, 1-\r+0.5)
        {\small\pgfmathprintnumber[fixed,fixed zerofill,precision=2]{\val}};
    \else
      \node[black] at (\c+0.5, 1-\r+0.5)
        {\small\pgfmathprintnumber[fixed,fixed zerofill,precision=2]{\val}};
    \fi
  }%
}%
\foreach \i/\p in {0/2, 1/1}{
  \node[anchor=east] at (-0, 1.5-\i) {\small $V_\p$};
}
\foreach \c/\cond in {0/{\textit{Base}}, 1/G, 2/R, 3/{R+G}, 4/{Few+R+G}}{
  \node[anchor=base] at (\c+0.5, 2.25) {\small \textbf{\cond}};
}
\foreach \c/\m in {0/Llama, 1/DeepSeek, 2/Mistral, 3/GPT, 4/Qwen}{
  \node[anchor=base] at (\c+0.5, -0.65) {\footnotesize \m};
}
\draw[thick] (0, 2) -- (5, 2);
\draw[thick] (0, 0) -- (5, 0);
\end{tikzpicture}
\caption{F$_1$ across condition (top header) and prompt variant (rows). $V_1$ denotes the unconstrained prompt (\S\ref{app:prompt_templates}); $V_2$ denotes the default single-unit prompt used throughout the main experiments.}
\label{fig:effect_of_prompt}
\end{figure}

\begin{table*}[!htbp]
\small
\centering
\begin{tabular}{@{}cccclrrrlrrrl@{}}
\toprule
\multirow{2}{*}{\textbf{Shot}} &
  \multirow{2}{*}{\textbf{Guided}} &
  \multirow{2}{*}{\textbf{Rubric}} &
  \multirow{2}{*}{\textbf{V}} &
  \multicolumn{1}{c}{\multirow{2}{*}{\textbf{Model}}} &
  \multicolumn{4}{c|}{\textbf{Single reference}} &
  \multicolumn{4}{c}{\textbf{Full corpus}} \\ \cmidrule(l){6-13} 
 &
   &
   &
   &
  \multicolumn{1}{c}{} &
  \multicolumn{1}{c}{P} &
  \multicolumn{1}{c}{R} &
  \multicolumn{1}{c}{F1} &
  \multicolumn{1}{c}{\#Seg.} &
  \multicolumn{1}{c}{P} &
  \multicolumn{1}{c}{R} &
  \multicolumn{1}{c}{F1} &
  \multicolumn{1}{c}{\#Seg.} \\ \midrule
\multirow{30}{*}{\textbf{Zero}} &
  \multirow{15}{*}{\textbf{\begin{tabular}[c]{@{}c@{}}Un-\\ guided\end{tabular}}} &
  \multirow{6}{*}{\textbf{\begin{tabular}[c]{@{}c@{}}No\\ Rubric\end{tabular}}} &
  \multirow{3}{*}{1} &
  Llama &
  0.061 &
  0.260 &
  0.099 &
  10309 &
  0.080 &
  0.232 &
  0.119 &
  15580 \\
 &  &                                  &                    & DeepSeek & 0.066 & 0.216 & 0.101 & 7743  & 0.086 & 0.190 & 0.118 & 11571 \\
 &  &                                  &                    & Mistral  & 0.073 & 0.276 & 0.115 & 9273  & 0.097 & 0.248 & 0.139 & 13779 \\ \cmidrule(l){4-13} 
 &  &                                  & \multirow{3}{*}{2} & Llama    & 0.145 & 0.216 & 0.174 & 3629  & 0.183 & 0.183 & 0.183 & 5363  \\
 &  &                                  &                    & DeepSeek & 0.136 & 0.227 & 0.170 & 4020  & 0.172 & 0.189 & 0.180 & 5851  \\
 &  &                                  &                    & Mistral  & 0.232 & 0.301 & 0.262 & 3161  & 0.251 & 0.221 & 0.235 & 4740  \\ \cmidrule(l){3-13} 
 &  & \multirow{9}{*}{\textbf{Rubric}} & \multirow{5}{*}{1} & Llama    & 0.058 & 0.267 & 0.095 & 11230 & 0.079 & 0.245 & 0.119 & 16763 \\
 &  &                                  &                    & DeepSeek & 0.068 & 0.239 & 0.106 & 8225  & 0.095 & 0.220 & 0.132 & 11924 \\
 &  &                                  &                    & Mistral  & 0.088 & 0.298 & 0.136 & 8227  & 0.123 & 0.280 & 0.171 & 12212 \\
 &  &                                  &                    & Qwen     & 0.093 & 0.270 & 0.138 & 7075  & 0.112 & 0.239 & 0.153 & 11448 \\
 &  &                                  &                    & GPT      & 0.195 & 0.354 & 0.252 & 4420  & 0.226 & 0.285 & 0.253 & 6764  \\ \cmidrule(l){4-13} 
 &  &                                  & \multirow{4}{*}{2} & Llama    & 0.190 & 0.253 & 0.217 & 3251  & 0.216 & 0.192 & 0.203 & 4766  \\
 &  &                                  &                    & Mistral  & 0.225 & 0.349 & 0.273 & 3781  & 0.241 & 0.261 & 0.251 & 5798  \\
 &  &                                  &                    & Qwen     & 0.302 & 0.346 & 0.322 & 2800  & 0.312 & 0.239 & 0.271 & 4112  \\
 &  &                                  &                    & GPT      & 0.344 & 0.357 & 0.350 & 2526  & 0.364 & 0.250 & 0.296 & 3681  \\ \cmidrule(l){2-13} 
 &
  \multirow{17}{*}{\textbf{Guided}} &
  \multirow{8}{*}{\textbf{\begin{tabular}[c]{@{}c@{}}No\\ Rubric\end{tabular}}} &
  \multirow{3}{*}{1} &
  Llama &
  0.057 &
  0.243 &
  0.093 &
  10351 &
  0.081 &
  0.234 &
  0.120 &
  15583 \\
 &  &                                  &                    & DeepSeek & 0.064 & 0.250 & 0.102 & 9054  & 0.087 & 0.228 & 0.125 & 13208 \\
 &  &                                  &                    & Mistral  & 0.094 & 0.306 & 0.144 & 7938  & 0.127 & 0.288 & 0.177 & 12127 \\ \cmidrule(l){4-13} 
 &  &                                  & \multirow{5}{*}{2} & Llama    & 0.279 & 0.354 & 0.312 & 3094  & 0.293 & 0.247 & 0.268 & 4529  \\
 &  &                                  &                    & DeepSeek & 0.196 & 0.316 & 0.242 & 3619  & 0.217 & 0.239 & 0.227 & 5326  \\
 &  &                                  &                    & Mistral  & 0.290 & 0.365 & 0.323 & 3069  & 0.301 & 0.259 & 0.278 & 4620  \\  
 &  &                                  &                    & Qwen     & 0.322 & 0.346 & 0.334 & 2623  & 0.341 & 0.245 & 0.285 & 3859  \\ 
 &  &                                  &                    & GPT      & 0.327 & 0.346 & 0.336 & 2573  & 0.347 & 0.246 & 0.288 & 3792  \\ \cmidrule(l){3-13} 
 &  & \multirow{9}{*}{\textbf{Rubric}} & \multirow{5}{*}{1} & Llama    & 0.052 & 0.249 & 0.087 & 11593 & 0.074 & 0.239 & 0.113 & 17316 \\
 &  &                                  &                    & DeepSeek & 0.061 & 0.240 & 0.097 & 8987  & 0.086 & 0.227 & 0.125 & 13467 \\
 &  &                                  &                    & Mistral  & 0.081 & 0.277 & 0.125 & 8359  & 0.114 & 0.263 & 0.159 & 12408 \\
 &  &                                  &                    & Qwen     & 0.115 & 0.303 & 0.167 & 6407  & 0.136 & 0.261 & 0.179 & 10305 \\
 &  &                                  &                    & GPT      & 0.072 & 0.315 & 0.118 & 10616 & 0.089 & 0.294 & 0.136 & 17739 \\ \cmidrule(l){4-13} 
 &  &                                  & \multirow{4}{*}{2} & Llama    & 0.256 & 0.334 & 0.290 & 3184  & 0.275 & 0.236 & 0.254 & 4592  \\
 &  &                                  &                    & Mistral  & 0.271 & 0.365 & 0.311 & 3286  & 0.288 & 0.267 & 0.277 & 4962  \\
 &  &                                  &                    & Qwen     & 0.373 & 0.403 & 0.387 & 2629  & 0.374 & 0.268 & 0.312 & 3838  \\
 &  &                                  &                    & GPT      & 0.356 & 0.387 & 0.371 & 2650  & 0.376 & 0.273 & 0.316 & 3892  \\ \bottomrule
\end{tabular}
\caption{\ourMethod{} results for all zero-shot configurations on the single reference subset and the full corpus, as defined in \S\ref{sec:experiments}. $V_1$ denotes the unconstrained prompt (Figure~\ref{fig:prompt-1-template}); $V_2$ denotes the default single-unit prompt (Figure~\ref{fig:prompt-2-template}), used throughout the main results. P and R denote precision and recall as formulated in \S\ref{sec:crossmatching}; \#Seg.\ denotes the number of generated segments. Model names are abbreviated for space; full names are given in \S\ref{sec:experiments}.}\label{tab:fullResultsZero}
\end{table*}
\begin{table*}[!htbp]
\small
\centering
\begin{tabular}{@{}cccclrrrlrrrl@{}}
\toprule
\multirow{2}{*}{\textbf{Shot}} &
  \multirow{2}{*}{\textbf{Guided}} &
  \multirow{2}{*}{\textbf{Rubric}} &
  \multirow{2}{*}{\textbf{V}} &
  \multicolumn{1}{c}{\multirow{2}{*}{\textbf{Model}}} &
  \multicolumn{4}{c|}{\textbf{Single reference}} &
  \multicolumn{4}{c}{\textbf{Full corpus}} \\ \cmidrule(l){6-13} 
 &
   &
   &
   &
  \multicolumn{1}{c}{} &
  \multicolumn{1}{c}{P} &
  \multicolumn{1}{c}{R} &
  \multicolumn{1}{c}{F1} &
  \multicolumn{1}{c}{\#Seg.} &
  \multicolumn{1}{c}{P} &
  \multicolumn{1}{c}{R} &
  \multicolumn{1}{c}{F1} &
  \multicolumn{1}{c}{\#Seg.} \\ \midrule
\multirow{30}{*}{\textbf{Few}} &
  \multirow{15}{*}{\textbf{\begin{tabular}[c]{@{}c@{}}Un-\\ Guided\end{tabular}}} &
  \multirow{6}{*}{\textbf{\begin{tabular}[c]{@{}c@{}}No\\ Rubric\end{tabular}}} &
  \multirow{3}{*}{1} &
  Llama &
  0.077 &
  0.231 &
  0.115 &
  7278 &
  0.102 &
  0.203 &
  0.136 &
  10653 \\
 &  &                                  &                    & DeepSeek & 0.070 & 0.237 & 0.108 & 7920  & 0.093 & 0.220 & 0.131 & 12097 \\
 &  &                                  &                    & Mistral  & 0.110 & 0.264 & 0.156 & 5819  & 0.148 & 0.241 & 0.183 & 8719  \\ \cmidrule(l){4-13} 
 &  &                                  & \multirow{3}{*}{2} & Llama    & 0.190 & 0.259 & 0.219 & 3327  & 0.218 & 0.198 & 0.208 & 4873  \\
 &  &                                  &                    & DeepSeek & 0.194 & 0.276 & 0.228 & 3419  & 0.221 & 0.214 & 0.218 & 5113  \\
 &  &                                  &                    & Mistral  & 0.274 & 0.318 & 0.294 & 2818  & 0.293 & 0.227 & 0.256 & 4151  \\ \cmidrule(l){3-13} 
 &  & \multirow{9}{*}{\textbf{Rubric}} & \multirow{5}{*}{1} & Llama    & 0.066 & 0.229 & 0.102 & 8461  & 0.089 & 0.201 & 0.124 & 12062 \\
 &  &                                  &                    & DeepSeek & 0.050 & 0.249 & 0.083 & 12142 & 0.068 & 0.229 & 0.105 & 17996 \\
 &  &                                  &                    & Mistral  & 0.100 & 0.282 & 0.147 & 6883  & 0.136 & 0.265 & 0.180 & 10440 \\
 &  &                                  &                    & Qwen     & 0.222 & 0.314 & 0.260 & 3443  & 0.257 & 0.249 & 0.253 & 5199  \\
 &  &                                  &                    & GPT      & 0.260 & 0.388 & 0.311 & 3641  & 0.299 & 0.310 & 0.305 & 5560  \\ \cmidrule(l){4-13} 
 &  &                                  & \multirow{4}{*}{2} & Llama    & 0.224 & 0.258 & 0.240 & 2808  & 0.235 & 0.180 & 0.204 & 4103  \\
 &  &                                  &                    & Mistral  & 0.241 & 0.344 & 0.283 & 3484  & 0.263 & 0.260 & 0.261 & 5305  \\
 &  &                                  &                    & Qwen     & 0.276 & 0.331 & 0.301 & 2919  & 0.292 & 0.234 & 0.260 & 4305  \\
 &  &                                  &                    & GPT      & 0.348 & 0.366 & 0.357 & 2557  & 0.369 & 0.256 & 0.302 & 3721  \\ \cmidrule(l){2-13} 
 &
  \multirow{15}{*}{\textbf{Guided}} &
  \multirow{6}{*}{\textbf{\begin{tabular}[c]{@{}c@{}}No\\ Rubric\end{tabular}}} &
  \multirow{3}{*}{1} &
  Llama &
  0.085 &
  0.238 &
  0.125 &
  6847 &
  0.117 &
  0.221 &
  0.153 &
  10123 \\
 &  &                                  &                    & DeepSeek & 0.050 & 0.263 & 0.084 & 12815 & 0.069 & 0.243 & 0.107 & 18947 \\
 &  &                                  &                    & Mistral  & 0.074 & 0.327 & 0.121 & 10720 & 0.103 & 0.303 & 0.154 & 15755 \\ \cmidrule(l){4-13} 
 &  &                                  & \multirow{3}{*}{2} & Llama    & 0.302 & 0.348 & 0.323 & 2812  & 0.304 & 0.231 & 0.262 & 4079  \\
 &  &                                  &                    & DeepSeek & 0.198 & 0.304 & 0.240 & 3442  & 0.220 & 0.234 & 0.227 & 5293  \\
 &  &                                  &                    & Mistral  & 0.295 & 0.350 & 0.320 & 2891  & 0.309 & 0.249 & 0.275 & 4316  \\ \cmidrule(l){3-13} 
 &  & \multirow{9}{*}{\textbf{Rubric}} & \multirow{5}{*}{1} & Llama    & 0.066 & 0.212 & 0.101 & 7777  & 0.093 & 0.194 & 0.126 & 11183 \\
 &  &                                  &                    & DeepSeek & 0.057 & 0.242 & 0.092 & 10286 & 0.081 & 0.228 & 0.119 & 15056 \\
 &  &                                  &                    & Mistral  & 0.098 & 0.305 & 0.149 & 7580  & 0.135 & 0.281 & 0.182 & 11198 \\
 &  &                                  &                    & Qwen     & 0.217 & 0.299 & 0.252 & 3338  & 0.251 & 0.237 & 0.244 & 5055  \\
 &  &                                  &                    & GPT      & 0.289 & 0.340 & 0.312 & 2867  & 0.318 & 0.259 & 0.285 & 4371  \\ \cmidrule(l){4-13} 
 &  &                                  & \multirow{4}{*}{2} & Llama    & 0.290 & 0.357 & 0.320 & 2993  & 0.302 & 0.245 & 0.271 & 4355  \\
 &  &                                  &                    & Mistral  & 0.253 & 0.341 & 0.290 & 3289  & 0.274 & 0.244 & 0.258 & 4774  \\
 &  &                                  &                    & Qwen     & 0.375 & 0.409 & 0.391 & 2657  & 0.377 & 0.274 & 0.317 & 3903  \\
 &  &                                  &                    & GPT      & 0.352 & 0.376 & 0.363 & 2604  & 0.369 & 0.263 & 0.307 & 3814  \\ \bottomrule
\end{tabular}
\caption{\ourMethod{} results for all few-shot (5-shot) configurations, evaluated on the same subsets and under the same prompt variants ($V_1$, $V_2$) as the zero-shot setting. Column definitions follow Table~\ref{tab:fullResultsZero}.}\label{tab:fullResultsFew}
\end{table*}
\begin{table*}[!htbp]
\small
\centering
\begin{tabular}{@{}cccclrrrlrrrl@{}}
\toprule
\multirow{2}{*}{\textbf{Shot}} &
  \multirow{2}{*}{\textbf{Guided}} &
  \multirow{2}{*}{\textbf{Rubric}} &
  \multirow{2}{*}{\textbf{V}} &
  \multicolumn{1}{c}{\multirow{2}{*}{\textbf{Model}}} &
  \multicolumn{4}{c|}{\textbf{Single reference}} &
  \multicolumn{4}{c}{\textbf{Full corpus}} \\ \cmidrule(l){6-13} 
 &
   &
   &
   &
  \multicolumn{1}{c}{} &
  \multicolumn{1}{c}{P} &
  \multicolumn{1}{c}{R} &
  \multicolumn{1}{c}{F1} &
  \multicolumn{1}{c}{\#Seg.} &
  \multicolumn{1}{c}{P} &
  \multicolumn{1}{c}{R} &
  \multicolumn{1}{c}{F1} &
  \multicolumn{1}{c}{\#Seg.} \\ \midrule
\multirow{12}{*}{\textbf{Zero}} &
  \multirow{6}{*}{\textbf{\begin{tabular}[c]{@{}c@{}}Un-\\ Guided\end{tabular}}} &
  \multirow{3}{*}{\textbf{\begin{tabular}[c]{@{}c@{}}No\\ Rubric\end{tabular}}} &
  \multirow{3}{*}{CoT} &
  Llama &
  0.114 &
  0.214 &
  0.149 &
  4464 &
  0.149 &
  0.187 &
  0.166 &
  6641 \\
 &  &                                  &                      & DeepSeek & 0.098 & 0.221 & 0.136 & 4708 & 0.129 & 0.201 & 0.157 & 7322 \\
 &  &                                  &                      & Mistral  & 0.149 & 0.285 & 0.196 & 4653 & 0.182 & 0.240 & 0.207 & 7065 \\ \cmidrule(l){3-13} 
 &  & \multirow{3}{*}{\textbf{Rubric}} & \multirow{3}{*}{CoT} & Llama    & 0.113 & 0.229 & 0.151 & 4857 & 0.145 & 0.195 & 0.167 & 7140 \\
 &  &                                  &                      & DeepSeek & 0.093 & 0.206 & 0.128 & 4325 & 0.130 & 0.197 & 0.157 & 6894 \\
 &  &                                  &                      & Mistral  & 0.137 & 0.308 & 0.189 & 5484 & 0.173 & 0.268 & 0.210 & 8312 \\ \cmidrule(l){2-13} 
 &
  \multirow{6}{*}{\textbf{Guided}} &
  \multirow{3}{*}{\textbf{\begin{tabular}[c]{@{}c@{}}No\\ Rubric\end{tabular}}} &
  \multirow{3}{*}{CoT} &
  Llama &
  0.113 &
  0.216 &
  0.148 &
  4615 &
  0.143 &
  0.184 &
  0.161 &
  6848 \\
 &  &                                  &                      & DeepSeek & 0.096 & 0.200 & 0.129 & 4271 & 0.127 & 0.181 & 0.149 & 6629 \\
 &  &                                  &                      & Mistral  & 0.137 & 0.275 & 0.183 & 4876 & 0.182 & 0.245 & 0.209 & 7214 \\ \cmidrule(l){3-13} 
 &  & \multirow{3}{*}{\textbf{Rubric}} & \multirow{3}{*}{CoT} & Llama    & 0.115 & 0.238 & 0.155 & 5001 & 0.146 & 0.198 & 0.168 & 7260 \\
 &  &                                  &                      & DeepSeek & 0.091 & 0.204 & 0.126 & 4534 & 0.128 & 0.195 & 0.154 & 7111 \\
 &  &                                  &                      & Mistral  & 0.127 & 0.294 & 0.178 & 5631 & 0.165 & 0.259 & 0.202 & 8431 \\ \bottomrule
\end{tabular}
\caption{\ourMethod{} results for chain-of-thought configurations under the unconstrained prompt ($V_1$). Column definitions follow Table~\ref{tab:fullResultsZero}.}\label{tab:fullResultsCoT}
\end{table*}

\section{Prompt Templates}\label{app:prompt_templates}
\begin{figure*}[p]
    \centering
    \includegraphics[width=0.95\textwidth,page=1]{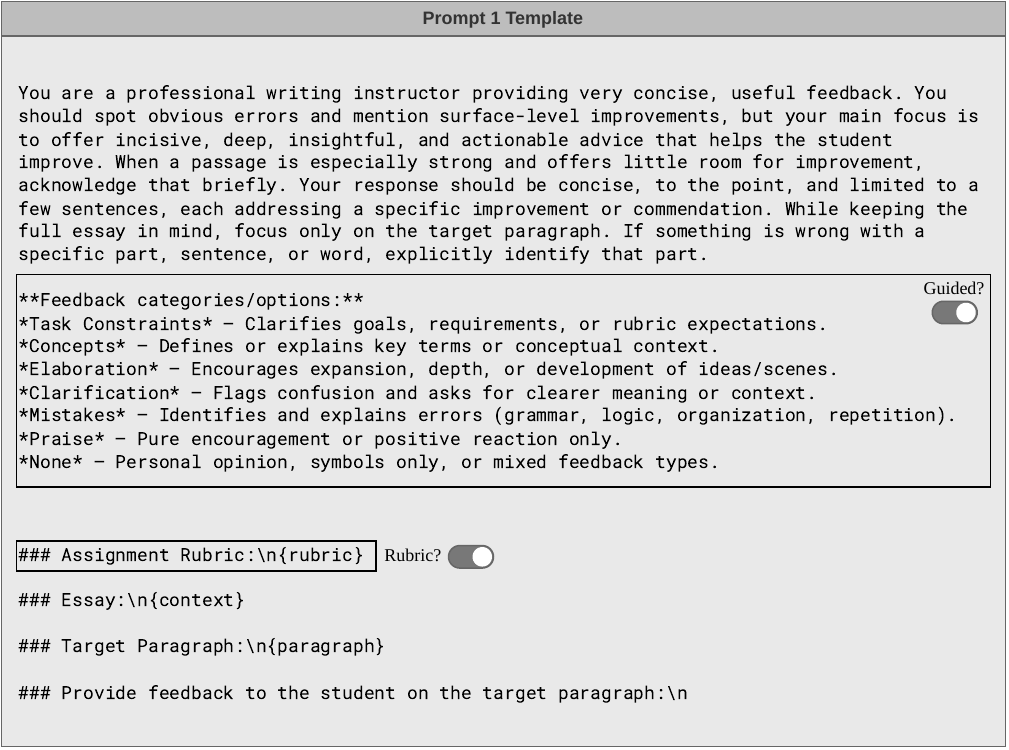}
\caption{Unconstrained prompt template (free-form paragraph-level feedback with no constraint on the number of feedback units). Used in the supplementary results comparing prompt designs (\S\ref{app:full_results}); the guided variant is introduced in \S\ref{sec:experiments}.}    \label{fig:prompt-1-template}
\end{figure*}

\begin{figure*}[p]
    \centering
    \includegraphics[width=0.95\textwidth,page=1]{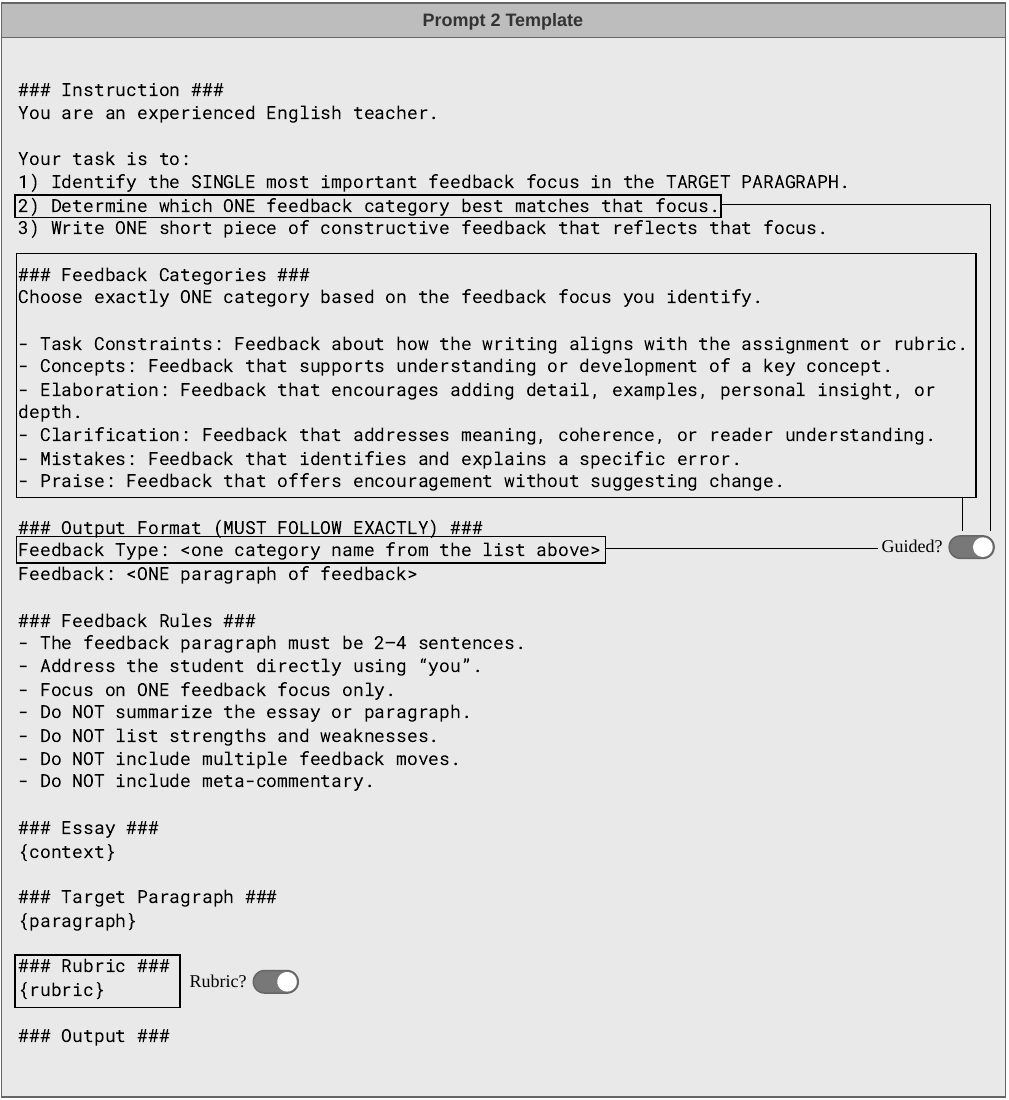}
\caption{Default single-unit prompt template used throughout the main experiments (\S\ref{sec:experiments}); the guided variant is also introduced \S\ref{sec:experiments}.}    \label{fig:prompt-2-template}
\end{figure*}

We use two prompt templates, both with two toggles (Figures~\ref{fig:prompt-1-template},~\ref{fig:prompt-2-template}): \emph{Guided?} controls whether the feedback categories (\S\ref{sec:experiments}) are listed, and \emph{Rubric?} whether the assignment prompt and rubric are inserted via \texttt{\{rubric\}}. The unconstrained template ($V_1$, Figure~\ref{fig:prompt-1-template}) places no limit on the number of feedback units; the constrained template ($V_2$, Figure~\ref{fig:prompt-2-template}) asks for a single most-important point under a fixed output format. $V_2$ is the default in the main experiments (\S\ref{sec:experiments}); $V_1$ is used only in the prompt-design comparison (\S\ref{app:full_results}).

\end{document}